\documentclass[journal]{IEEEtran}
\usepackage{cite}

\ifCLASSINFOpdf
\else
\fi
\usepackage{amsmath}
\usepackage{amssymb}
\usepackage[pdftex]{graphicx}
\usepackage{algorithmic}
\usepackage{url}
\usepackage{color}

\begin{document}
%
\title{Elevation Estimation-Driven Building 3D Reconstruction from Single-View Remote Sensing Imagery}
%
%
%

\author{Yongqiang Mao,
        Kaiqiang Chen,
        Liangjin Zhao,
        Wei Chen,\\
        Deke Tang,
        Wenjie Liu,
        Zhirui Wang,
        Wenhui Diao,
        Xian Sun, 
        Kun Fu
\thanks{This work was supported by National Key R\&D Program of China 
under Grant No. 2021YFB3900504. (Corresponding author: Kaiqiang Chen.)

Yongqiang Mao, Wenjie Liu, Xian Sun, and Kun Fu are with the Aerospace Information Research Institute, Chinese Academy of Sciences, Beijing 100190, China, the Key Laboratory of Network Information System Technology (NIST), Aerospace Information Research Institute, Chinese Academy of Sciences, Beijing 100190, China, the University of Chinese Academy of Sciences and the School of Electronic, Electrical and Communication Engineering, University of Chinese Academy of Sciences,
Beijing 100190, China (e-mail: maoyongqiang19@mails.ucas.ac.cn; liuwenjie18@mails.ucas.ac.cn; sunxian@aircas.ac.cn; kunfuiecas@gmail.com).

Kaiqiang Chen, Liangjin Zhao, Zhirui Wang, and Wenhui Diao are with the Aerospace Information Research Institute, Chinese Academy of Sciences, Beijing 100190, China and the Key Laboratory of Network Information System Technology (NIST), Aerospace Information Research Institute, Chinese Academy of Sciences, Beijing 100190, China (e-mail: chenkaiqiang14@mails.ucas.ac.cn;).

Wei Chen and Deke Tang are with the Geovis Technology Co., Ltd. (e-mail: chenwei@geovis.com.cn; tangdk@geovis.com.cn)

}
}

%
%

\markboth{Journal of \LaTeX\ Class Files,~Vol.~14, No.~8, August~2015}%
{Shell \MakeLowercase{\textit{et al.}}: Bare Demo of IEEEtran.cls for IEEE Journals}
%



\maketitle



\begin{abstract}
Building 3D reconstruction from remote sensing images has a wide range of applications in smart cities, photogrammetry and other fields. Methods for automatic 3D urban building modeling typically employ multi-view images as input to algorithms to recover point clouds and 3D models of buildings. However, such models rely heavily on multi-view images of buildings, which are time-intensive and limit the applicability and practicality of the models. To solve these issues, we focus on designing an efficient DSM estimation-driven reconstruction framework (Building3D), which aims to reconstruct 3D building models from the input single-view remote sensing image. Existing DSM estimation networks suffer from the imbalance between local features and global features, which leads to over-smooth DSM estimates at instance boundaries. To address this issue, we propose a Semantic Flow Field-guided DSM Estimation (SFFDE) network, which utilizes the proposed concept of elevation semantic flow to achieve the registration of local and global features. First, in order to make the network semantics globally aware, we propose an Elevation Semantic Globalization (ESG) module to realize the semantic globalization of instances. Further, in order to alleviate the semantic span of global features and original local features, we propose a Local-to-Global Elevation Semantic Registration (L2G-ESR) module based on elevation semantic flow. Our Building3D is rooted in the SFFDE network for building elevation prediction, synchronized with a building extraction network for building masks, and then sequentially performs point cloud reconstruction, surface reconstruction (or CityGML model reconstruction). On this basis, our Building3D can optionally generate CityGML models or surface mesh models of the buildings. Extensive experiments on ISPRS Vaihingen and DFC2019 datasets on the DSM estimation task show that our SFFDE significantly improves upon state-of-the-arts and $\delta_1$, $\delta_2$ and $\delta_3$ metrics of our SFFDE are improved to 0.595, 0.897 and 0.970. Furthermore, our Building3D achieves impressive results in the 3D point cloud and 3D model reconstruction process.
\end{abstract}


%
\IEEEpeerreviewmaketitle

\section{Introduction}
\IEEEPARstart{T}{hanks} to the increasing development of various sensors, various remote sensing data have been applied in different fields~\cite{hu2022pseudo,li2022a3clnn}, including semantic segmentation~\cite{mou2020relation, deng2021ccanet} and object detection~\cite{wei2020x, sun2021pbnet, wei2020oriented} of 2D remote sensing image, and 3D point cloud classification~\cite{qi2017pointnet, mao2022beyond, mao2022semantic}. Among them, the interpretation of interactive information between two-dimensional data and three-dimensional data gradually enters researchers' field of vision. 3D reconstruction of urban buildings is a significant constituent of remote sensing image interpretation, where the goal is to parse image or point cloud data to generate 3D model representations~\cite{kim20023d,kuschk2013model,ozcanli2014automatic,wang2016efficient,wu2014building,hepp2018plan3d} of urban buildings. In recent years, large-scale geospatial mesh models of urban buildings have been introduced widespreadly in many fields, such as navigation and urban planning~\cite{mittal2019vision}, urban 3D maps~\cite{hajek2016principles}, etc. 

\begin{figure*}
    \centering
    \includegraphics[width=1.0\linewidth]{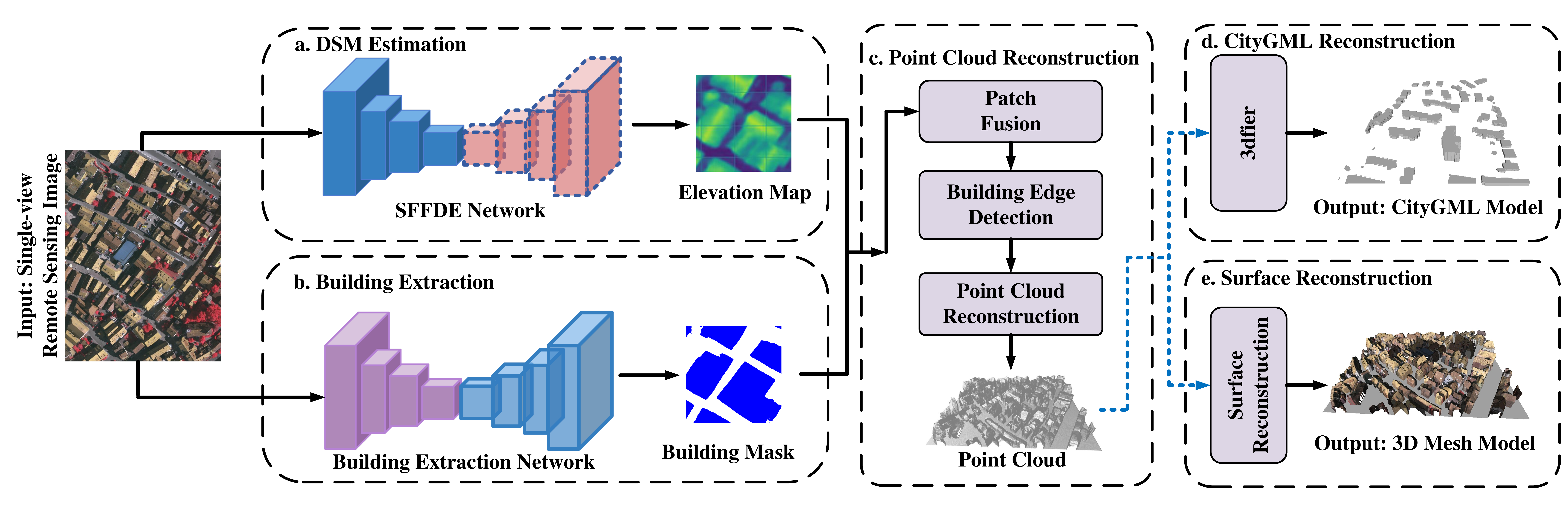}
    \caption{Flowchart of the proposed Building3D framework for 3D reconstruction of buildings. The input of Building3D is a single-view remote sensing image, the outputs are the corresponding elevation maps, building masks, point cloud, and 3D building model. The blue dotted line represents optional follow-up operations (CityGML models or Mesh models).}
    \label{motivation}
\end{figure*}

Among the existing 3D model acquisition methods, there are mainly four approaches: airborne lidars, Geiger-mode lidars, reconstruction based on multi-view UAV images, and reconstruction based on single-view remote sensing images. Although airborne lidar can obtain accurate 3D point clouds ~\cite{Cramer_2010, Rottensteiner_et_al_2012, niemeyer2014contextual,ye2020lasdu,le20192019, bosch2019semantic} of buildings in the target area, it is limited by high acquisition cost and low acquisition efficiency and difficult to apply to large-scale tasks. Recent Geiger-mode lidars ~\cite{kim2013simulation} can acquire data efficiently, they suffer from high noise and low penetration. Building reconstruction based on multi-view UAV images ~\cite{kim20023d,ozcanli2014automatic} can acquire 2D image data for 3D model reconstruction at low cost, but it has a very high time cost for large-scale regional reconstruction and requires a high degree of overlap between images. Unlike airborne lidar and drone images, single-view remote sensing imagery can cover a larger area with high stability and low cost. Therefore, we focus on the 3D reconstruction of buildings based on single-view remote sensing imagery (Fig.~\ref{motivation}), aiming to utilize low-cost large-scale remote sensing imagery for fast urban building reconstruction.

For single-view-based building 3D reconstruction methods, some methods~\cite{yu2021automatic} derive 3D building models (e.g., CityGML) and only focus on the restoration of roof structure topology. The researchers aim to reconstruct the roof topology using traditional methods, and then obtain 3D data in the form of CityGML based on the building height information. Some researchers use pre-labeled roof structure types to train the network and then extract the roof topology. However, these methods are limited to the extraction of topology structure, which brings unnecessary extraction of artificial prior knowledge for building reconstruction, resulting in cumbersome and redundant reconstruction process, which is not conducive to building reconstruction in large-scale areas. In order to solve this problem, based on the prediction of building elevation information, we perform 3D mapping of building elevation to obtain building point cloud data, and then selectively reconstruct the subsequent CityGML model or mesh model of the buildings. Based on this, our reconstruction framework (Fig. ~\ref{motivation}) can simultaneously reconstruct more realistic mesh models of buildings and virtual models like CityGML.

For single-view remote sensing images, the elevation information of buildings is a key factor in constructing the 3D building models. Although substantial progress has been made in existing elevation prediction methods~\cite{li2020geometry, mou2018im2height, mahdi2020aerial, wang2020boundary,batra2012learning, saxena2005learning} from remote sensing images, the problem of low prediction accuracy occurs when both regular and irregular textured objects are present. Through our investigation, we realize that the low prediction accuracy of textured regular and irregular objects is caused by the inability of the receptive field to achieve a local-global trade-off. The imbalance between global and local features will cause the network to suffer from insufficient receptive field in the learning process, and it cannot take into account the extraction of regular texture and irregular texture object features at the same time. Since the features extracted by CNN have local perception, to solve this problem, we introduce elevation semantic globalization to obtain global feature perception. However, the network at this time does not have the ability to take into account and trade off local and global features at the same time. To address this issue, we introduce the concept of elevation semantic flow embedded in a local-to-global feature registration module to explicitly model the changing trend between global and local features to trade off local and global features. Based on this, we propose an Elevation Semantic Flow Field-guided DSM Estimation (SFFDE) network (Fig. ~\ref{motivation}) to obtain accurate elevation predictions for various textured objects.

In this paper, we propose a single-view remote sensing image-based building reconstruction framework (Building3D) based on the SFFDE network, as illustrated in Fig. ~\ref{motivation}. Unlike previous research schemes targeting roof topology, our Building3D aims to recover both the CityGML model as well as the real mesh model of the buildings. Our Building3D framework consists of four parts: Semantic Flow Field-guided DSM Estimation (SFFDE) network, building extraction network, point cloud reconstruction process, and building reconstruction process. Given a single-view remote sensing image, the elevation information of the image is first predicted through our SFFDE network, which is mainly composed of two parts: Elevation Semantic Globalization (ESG) and Local-to-Global Elevation Semantic Registration (L2G-ESR), which realize the purpose of feature globalization and local and global feature registration, respectively. At the same time, the building mask is extracted from the image through the building extraction network. Next, the building elevation is extracted using the building mask, and the elevations of other uninteresting objects are filtered out. Based on this, the building elevation is used to reconstruct the point cloud data to obtain the 3D point cloud of the building. Finally, CityGML reconstruction or surface reconstruction is performed using the point cloud to obtain the final CityGML or 3D mesh model.

The contributions of this study are summarized as follows:
\begin{itemize}
    \item We introduce a Semantic Flow Field-guided DSM Estimation (SFFDE) network based on the proposed elevation semantic flow field for the local-to-global registration of elevation semantics. Specifically, Elevation Semantic Globalization (ESG) and Local-to-Global Elevation Semantic Registration (L2G-ESR) are designed to achieve the globalization and local-to-global registration of features, respectively.
    
    \item We propose a single-view image based building 3D reconstruction framework (Building3D) rooted in the proposed SFFDE network and extract building masks according to the building extraction network for subsequent 3D point cloud and building reconstruction.
    
    \item Experiments on ISPRS Vaihingen~\cite{vaihingen.org} and DFC2019~\cite{bosch2019semantic} datasets show that our SFFDE network significantly outperforms the baseline model, achieving new state-of-the-art. What's more, our Building3D achieves impressive results in the 3D point cloud and building reconstruction process. 
\end{itemize}

The remainder of this article is organized as follows: In Section II, we give a brief review of the related works on DSM estimation, building extraction, and building reconstruction. The details of our proposed framework Building3D and semantic flow field-guided DSM estimation (SFFDE) network are given in Section III. In Section IV, extensive experiments are conducted on ISPRS Vaihingen dataset and DFC2019 dataset to demonstrate the effectiveness of our SFFDE network. Also, the 3D reconstruction experiments of buildings are given in Section IV. Finally, Section V concludes this article.

\section{Related Work}
\subsection{DSM Estimation}
The existing methods ~\cite{li2020geometry, mou2018im2height, mahdi2020aerial, wang2020boundary} for DSM elevation estimation in the remote sensing field are mainly divided into three types: random field-based methods, CNN-based methods, and some hybrid methods. In random field-based methods, researchers ~\cite{batra2012learning, saxena2005learning} utilize conditional random field (CRF) and Markov random field (MRF) to model the local and global structure of an image. Considering that local features cannot provide sufficient features for predicting depth values, Batra and Saxena~\cite{batra2012learning} model the relationship between adjacent regions. Furthermore, in order to obtain global features beyond the local ones, Saxena et al.~\cite{saxena2005learning,saxena20083} first compute the features of the four nearest neighbors of the specified patch, and then use the MRF and Laplacian models to estimate the depth of each patch. In recent years, CNN has been widely used in the fields of object classification, semantic segmentation, and object detection. Inspired by this, some researchers~\cite{amirkolaee2019height} propose a ResNet-based convolutional network to predict DSM elevation information. In IMG2DSM~\cite{ghamisi2018img2dsm}, an adversarial loss function is introduced to improve the possibility of synthesizing DSM. It also employs a conditional generative adversarial network to build image-to-DSM elevation translation. Zhang and Chen~\cite{zhang2019multi} improve the learning ability of abstract features of objects of different scales by means of multi-scale feature extraction through a multi-path fusion network. 
Li et al.~\cite{li2020height} divided the height values into intervals with increasing spacing, and transformed the regression problem into an ordinal regression problem, using an ordinal loss for network training. After that, a post-processing technique is designed to convert the predicted height map of each patch into a seamless height map. Carvalho et al.~\cite{carvalho2018regression} have studied various loss functions of depth regression in depth, and combined the encoder-decoder architecture with adversarial loss, and then proposed a new depth estimation network D3-Net.
In the hybrid methods, Wang et al.~\cite{wang2015towards} propose a joint framework incorporating hierarchical conditional random fields, aiming to predict depth and segmentation outcomes from a single-view remote sensing image. Srivastava et al.~\cite{srivastava2017joint} use a method for supervised network training via a multi-task loss and introduce a unified framework for elevation estimation and semantic segmentation of single-view remote sensing image.

However, none of these methods take into account the need for accurate elevation prediction to simultaneously guarantee the extraction of representative features for both regular and irregular textured objects, which is caused by the imbalance between global and local features. To address this issue, we propose a semantic flow field-guided DSM estimation network to trade off local and global features.

\subsection{Building Extraction}
Many researchers ~\cite{lin1998building,baatz1999object,wang2005building} have devoted themselves to designing efficient methods for automatically extracting buildings, which mainly consists of methods based on handcrafted features and methods based on deep learning. In the handcrafted feature-based methods, geometric, spectral, and contextual information of buildings are used to design representative features for accurate building extraction. Lin and Nevatia ~\cite{lin1998building} first apply edge detection to building extraction for the detection of roofs, walls and shadows. Later, the fractal network algorithm is proposed by Baatz ~\cite{baatz1999object} to segment the image at multiple scales and extract the target building by combining the texture and other features of the image. Wang and Liu~\cite{wang2005building} achieve the purpose of pixel-by-pixel classification of images by using machine learning methods to receive input from feature vectors constructed from image texture, shape, and structural features. However, methods based on handcrafted features often only utilize the shallow features of objects, ignoring the semantics of deep features. Therefore, methods based on deep learning gradually enter the field of vision of researchers. Minh~\cite{mnih2013machine} applies an image patch-based building block extraction method to building extraction, an early work on convolutional neural networks applied to building extraction. On this basis, Minh also uses CRF or post-processing to refine the results of the network. Later, Huang et al.~\cite{huang2016building} introduce an improved DeconvNet that adds upsampling and cheat connection operations to deconvolution layers to achieve building extraction. Wu et al.~\cite{wu2018automatic} propose a multi-constraint FCN to perform feature learning on remote sensing images for the purpose of automatically extracting buildings.

In this paper, we apply the popular segmentation network DeeplabV3+~\cite{chen2018encoder} to generate binary masks for buildings.

\begin{figure*}
    \begin{center}
    \includegraphics[width=1.0\linewidth]{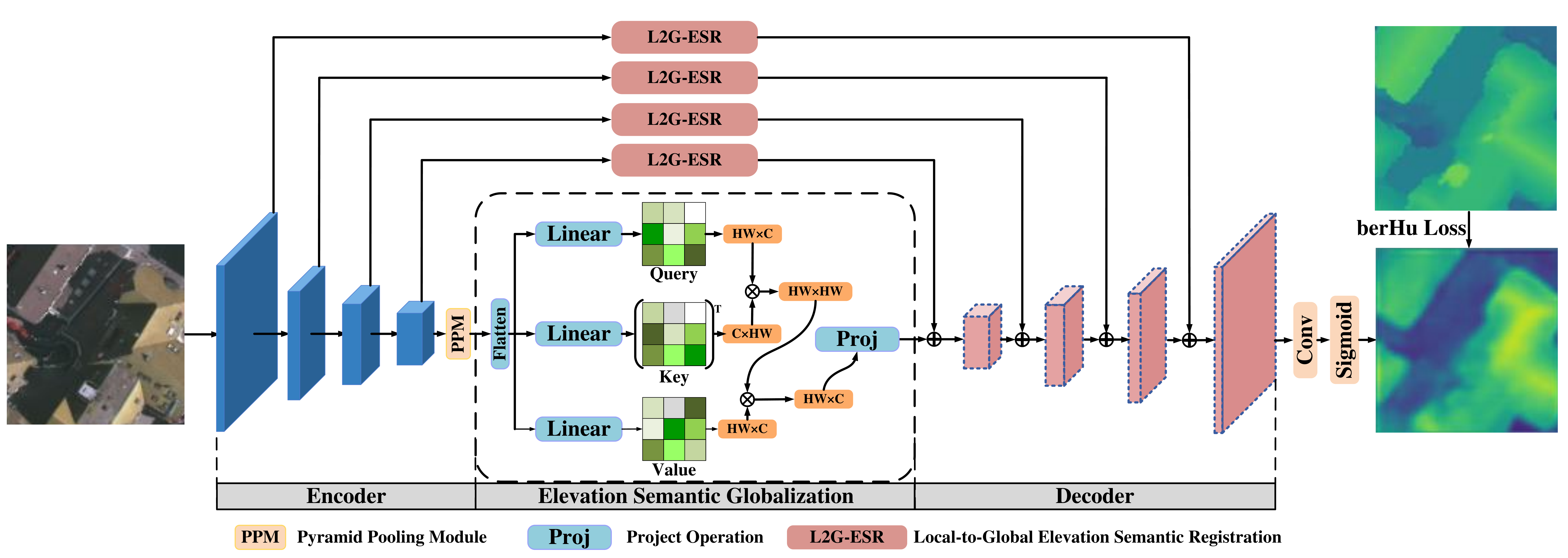}
    \end{center}
    \setlength{\abovecaptionskip}{0pt}
    \caption{Flowchart of the proposed Semantic Flow Field-guided DSM Estimation (SFFDE) network for DSM estimation. The proposed elevation semantic globalization (ESG) is responsible for the global semantic representation. Furthermore, local-to-global elevation semantic registration (L2G-ESR) is introduced to achieve the trade-off of local and global features.}
    \label{SFFDE}
\end{figure*}

\subsection{Building Reconstruction}
The existing methods for 3D building reconstruction mainly focus on the roof topology restoration. Researchers~\cite{bulatov2014context,li2016reconstructing,yan2017hierarchical,alidoost20192d} extract roof vertices, eave lines, or segmented planes as primitives from an image or DSM through image segmentation, edge detection, plane patch detection, etc. After that, the combination, segmentation, topologically analysis of the individual geometric primitives are executed in sequence to generate the 3D shapes of the buildings. Bulatov et al.~\cite{bulatov2014context} extract building regions by performing vegetation and outlier filtering on a normalized DSM (nDSM), and then combine the extracted building pixels with operations such as graph-based orthophoto segmentation,  dominant direction extraction, polygonization, and generalization to extract buildings contour. Finally, they construct the building model by topologically analyzing the roof details. Li et al.~\cite{li2016reconstructing} classify MVS point clouds through graph-cut based Markov random fields (MRF), followed by RANSAC and regularized MRF for optimization and roof structure extraction, respectively.

However, these methods suffer from the limitations of multi-view image input or surfel geometric topology, thus bringing great difficulties to building reconstruction in large scenes. To address this issue, we propose a DSM estimation-driven framework for building 3D reconstruction with single-view remote sensing image as input.

\section{Our Approach}

\subsection{Overview}
The flowchart of our proposed building 3D reconstruction framework (Building3D) is illustrated in Fig.~\ref{motivation}. As a single-view image-based building 3D reconstruction approach, our framework consists of four stages: DSM estimation branch, building extraction branch, point cloud reconstruction branch and building reconstruction branch (surface reconstruction or CityGML reconstruction). Given an input remote sensing image, the DSM estimation branch and the building extraction branch first come into play to estimate the elevation information and extract the buildings, respectively. On this basis, the information obtained by the building extraction branch is used to filter the information from the elevation results predicted in the DSM estimation branch to obtain only the elevation information of buildings. Next, the point cloud data corresponding to the building is recovered according to the input image and the predicted elevation information. Finally, surface reconstruction or CityGML reconstruction is performed on the point cloud of buildings obtained by the point cloud reconstruction branch to obtain the final mesh or CityGML model. In the following, we first introduce the semantic flow field-driven DSM estimation (SFFDE) network, and then introduce the framework's building extraction process, point cloud reconstruction process, and building reconstruction process.

\subsection{Semantic Flow Field-guided DSM Estimation Network}
Considering generating fine 3D point clouds, the accurate DSM information is necessary. To address this issue, we design a Semantic Flow Field-guided DSM Estimation (SFFDE) network, as shown in Fig.~\ref{SFFDE}. Our SFFDE selects PSPNet~\cite{zhao2017pyramid} as the baseline and consists of three stages: First, a single remote sensing image is sent into a feature extraction network (such as resnet101) and a PPM (Pyramid Pooling Module)~\cite{zhao2017pyramid} module to obtain high-level representations of features. Second, an Elevation Semantic Globalization (ESG) module is introduced for the globalization of semantic features to obtain high-level global semantic representation. Finally, we propose a Local-to-Global Elevation Semantic Registration (L2G-ESR) module to achieve the registration of low-level local semantics and high-level global semantics of features between different resolutions. Furthermore, our SFFDE network is driven by the berHuLoss~\cite{zwald2012berhu} during training.

\subsubsection{Elevation Semantic Globalization}
The stacking of convolutional layers and pooling layers can increase the receptive field, but the receptive field of the convolution kernel of a specific layer on the original image is limited, which is unavoidable in local operations. However, elevation estimation based on single view remote sensing image requires more information on the original image. If the global information can be introduced in some layers, the problem that the local operation of the convolution cannot capture the global information can be well solved, and it can bring richer information to the subsequent layers. Inspired by the transformer\cite{vaswani2017attention} architecture, we introduce the Elevation Semantic Globalization (ESG) module (Fig.~\ref{SFFDE}) for the global semantic representation. 

Given feature maps $\textbf{F} \in \mathbb{R}^{H\times W\times C}$($H, W, C$ stand for height, width, and number of channels, respectively.), after the feature extraction of the backbone, these feature maps are high-level semantic features with local perception. Before being sent to the ESG module, the feature map $\textbf{F}$ is stretched from a 2D raster image into a 1D vector (regardless of the channel dimension), which can be formulated as:
\begin{equation}
    \mathcal{F} = flatten\left(\textbf{F}\right)
\end{equation}
where $\mathcal{F} \in \mathbb{R}^{N\times C}$($N=H\times W$). The stretched feature $\mathcal{F}$ goes through three fully connected layers in parallel for feature transformation, so that the 2D raster image features are mapped to the same space as the 1D vector features, which is formulated as:
\begin{equation}
\begin{aligned}
    \Big\{\mathbf{Q}, \mathbf{K}, \mathbf{V}\Big\} &= \Big\{FC_1(\mathcal{F}), FC_2(\mathcal{F}), FC_3(\mathcal{F})\Big\} \\
    &=\Big\{\mathbf{W}_q\mathcal{F}, \mathbf{W}_k\mathcal{F}, \mathbf{W}_v\mathcal{F}\Big\}
\end{aligned}    
\end{equation}
where $\mathbf{Q}, \mathbf{K}, \mathbf{V} \in \mathbb{R}^{N\times D}$. $D$ is the number of channels of the output features. Given query embeddings $\mathbf{Q}uery$ to be enhanced and feature maps $\mathbf{K}ey, \mathbf{V}alue$ to be fused, the ESG operation is defined as
\begin{equation}
    \mathbf{F}_{esg}=ESG\big(\mathbf{Q},\mathbf{K},\mathbf{V}\big) = Softmax\Big(\mathbf{W}_q\mathcal{F}{(\mathbf{W}_k\mathcal{F})}^{\top }\Big)\mathbf{W}_v\mathcal{F}
\end{equation}
where $ \mathbf{F}_{esg} \in \mathbb{R}^{N\times D}$ is the query feature after ESG operation.

After this, we set up a multi-head ESG operation module based on ESG operation and transformer architecture, the formula is as follows:
\begin{equation}
    \mathbf{F}_{out} = MHE(\mathbf{W}_q\mathcal{F}, \mathbf{W}_k\mathcal{F}, \mathbf{W}_v\mathcal{F})
\end{equation}
where $\mathbf{F}_{out}$ is the output features of MHE. MHE is the multi-head ESG operation which is similar to multi-head attention (MHA) operation~\cite{vaswani2017attention}. Specifically, each ESG head can be expressed as:
\begin{equation}
    Head_i=ESG(Q,K,V)
\end{equation}
where $Q,K,V$ represents the \textbf{Q}uery, \textbf{K}ey, and \textbf{V}alue matrix, respectively. Then, based on this, multi-head ESG can be expressed mathematically as:
\begin{equation}
    MultiHead(Q,K,V)=Cat(Head_1,…,Head_8)W
\end{equation}
where $Cat$ is the concatenate operation and $W$ represent the weight of the final full connection operation. Different ESG heads represent different subspaces, and the results of all ESG heads are spliced together to obtain the final result through full connection.

Next, a feature projection operation is adopted to perform feature mapping between spatial one-dimensional features and spatial two-dimensional features. Formally, the obtained projection feature $\mathbf{F}_{proj}$ can be defined as:
\begin{equation}
    \mathbf{F}_{proj} = Relu\Big(LN\big(FC(\mathbf{F}_{out})\big)\Big)
\end{equation}
where LN is the LayerNorm~\cite{ba2016layer} operation. Finally, the projection feature $\mathbf{F}_{proj} \in \mathbb{R}^{N\times D}$ is reshaped to the same resolution as $\mathbf{F}$.
\subsubsection{Elevation Semantic Flow}
To register features at different resolutions, inspired by Optical Flow, we introduce the concept of Elevation Semantic Flow (ESF). The Optical Flow is the method that employs the changes of pixels in the image sequence in the time domain and the correlation between adjacent frames to find the corresponding relationship between the previous frame and the current frame, so as to calculate the motion of objects between adjacent frames. Analogous to the instantaneous gray change rate of pixels at the same location between video frames in a short period of time, we define a semantic change rate of elevation, that is, the semantic displacement of pixels between adjacent resolution images, which can be expressed as:
\begin{equation}
\begin{aligned}
    \vec{u} =(\frac{\partial x}{\partial l}, \frac{\partial y}{\partial l})
\end{aligned}
\end{equation}
where $l$ represents the varying resolution of feature maps and $\vec{u}$ is the elevation semantic flow vector, which represents the rate of change of semantics along the $x$ and $y$ directions.

In space, the motion can be described by a motion field. In the video, the motion of an object is often represented by an optical flow field. In the feature map of the elevation estimation network, the semantic motion can be defined by the feature vectors represented by the pixels of the feature maps of different resolutions. Thus, we propose to define an elevation semantic field of $F$, which is formulated as:
\begin{equation}
\begin{aligned}
    -\frac{\partial F}{\partial l} = \nabla  F \cdot\vec{u} = \frac{\partial F}{\partial x} \frac{\partial x}{\partial l} + \frac{\partial F}{\partial y}\frac{\partial y}{\partial l}   
\end{aligned}
\end{equation}
Similar to the optical flow field, the elevation semantic field is a two-dimensional vector field, which reflects the semantic change trend of each point in the feature map. 
Classical optical flow estimation is solved by a linear algebra method (such as the LK algorithm~\cite{lucas1981iterative}). We use an end-to-end training nonlinear optimization method and use Local-to-Global Elevation Semantic Registration (L2G-ESR) as a constraint to learn the elevation semantic flow field.

\begin{figure*}[t]
    \begin{center}
    \includegraphics[width=0.8\linewidth]{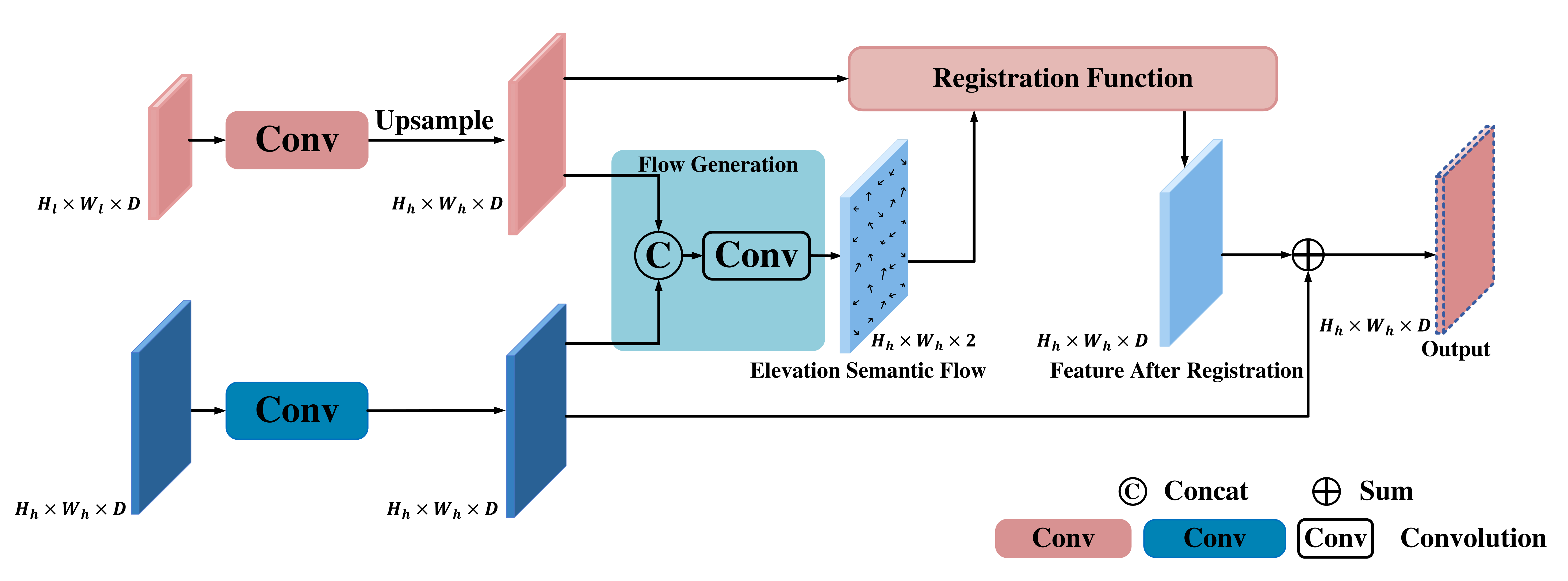}
    \end{center}
    \setlength{\abovecaptionskip}{0pt}
    \caption{Overview of our local-to-global elevation semantic registration (L2G-ESR) module. The inputs to L2G-ESR are two features of different resolutions. Flow generation and feature registration are performed sequentially. The pink and blue convolution blocks in the figure represent feature mapping operations for low-resolution and high-resolution, respectively. Specifically, both convolution operations are implemented using 1x1 two-dimensional convolution, and the input and output channel dimensions are the same.}
\label{L2G-ESR}
\end{figure*}
\subsubsection{Local-to-Global Elevation Semantic Registration}
After the globalization of features, the perceptual ability of features is extended from local to global. This enables the network to not only perceive low-dimensional local features, but also acquire the ability to perceive high-dimensional global features. However, since low-dimensional local features and global high-dimensional features are obtained by local convolution and elevation semantic globalization (not convolution operations) respectively, there is a huge span in the range of feature perception between global features and local features. In addition, these features do not have both local and global perception capabilities at the same time. To address these issues, local-to-global elevation semantic registration (L2G-ESR) is introduced, as shown in Fig.~\ref{L2G-ESR}.

Given the high-resolution local feature $\textbf{F}_h$ and the low-resolution globalized feature $\textbf{F}_l$, we first map $\textbf{F}_h$ and $\textbf{F}_l$ to the same number of channels as:
\begin{equation}
    \Big\{\widehat{\textbf{F}}_h, \widehat{\textbf{F}}_l\Big\} = \Big\{Conv_1(\textbf{F}_h), Conv_2(\textbf{F}_l)\Big\}
\end{equation}
where $\widehat{\textbf{F}}_h\in \mathbb{R}^{H_h\times W_h\times D}$ and $\widehat{\textbf{F}}_l\in \mathbb{R}^{H_l\times W_l\times D}$ are the new features after mapping.
$Conv_1$ and $Conv_2$ correspond to the blue and pink convolution blocks in the figure, respectively. Specifically, both convolution operations are implemented using 1x1 two-dimensional convolution, and the input and output channel dimensions are the same.
Then, we bilinearly interpolate the new low-resolution features $\widehat{\textbf{F}}_l$ to the same resolution as the high-resolution features $\widehat{\textbf{F}}_h$. On this basis, the two same-resolution features are fused and further encoded into a 2D elevation semantic flow field. Mathematically, the elevation semantic flow field $\textbf{S}$ can be expressed as:
\begin{equation}
\begin{aligned}
    \textbf{S} = Conv\Big(cat\big(upsample(\widehat{\textbf{F}}_l), \widehat{\textbf{F}}_h\big)\Big)
\end{aligned}
\end{equation}
where $upsample$ represents the bi-linear interpolation operation, $cat$ is the concatenation operation, and $Conv$ denotes the convolution operation. The semantic flow field $\textbf{S}\in \mathbb{R}^{H_h\times W_h\times 2}$ represents the change trend of elevation semantics between different resolutions. The channel `2' refers to the change of semantics in both $x$ and $y$ directions.

Let $\mathbf{L}\in \mathbb{R}^{H_h\times H_h\times 2}$ denote the coordinates of each pixel in the feature map $\widehat{\mathbf{F}}_h$. Then the generated semantic flow field $\mathbf{S}$ is employed to get the coordinates $\widehat{\mathbf{L}}$ of each pixel of the feature map after offset, as $\widehat{\mathbf{L}} = \mathbf{L} + \mathbf{S} (x'=x+\Delta x, y'=y+\Delta y)$.

Following the bi-linear sampling method in STN~\cite{jaderberg2015spatial}, the pixel $\mathbf{F}_{reg;ab}$ at $(a, b)$ of the output feature $\mathbf{F}_{reg}$ after feature registration operation is defined as:
\begin{equation}
\begin{aligned}
    &\mathbf{F}_{reg;ab} = R(\widehat{\mathbf{F}}_l)\\
    &=\sum_{m=1}^{H}\sum_{n=1}^{W}upsample(\widehat{\mathbf{F}}_{l})_{mn}\cdot max(0, 1-|\widehat{\mathbf{L}}_{x;ab}-m|)\\
    &\cdot max(0, 1-|\widehat{\mathbf{L}}_{y;ab}-n|)
\end{aligned}
\end{equation}
where $upsample(\widehat{\mathbf{F}}_{l})_{mn}$ is the pixel at $(m,n)$ of the aforementioned $\widehat{\mathbf{F}}_l$ after upsampling. Furthermore, $\widehat{\mathbf{L}}_{x;ab}$ and $\widehat{\mathbf{L}}_{y;ab}$ are the $x$ and $y$ coordinates of each pixel of $\widehat{\mathbf{L}}$. After that, the feature $\mathbf{F}_{reg}$ after registration is fused with aforementioned $\widehat{\mathbf{F}}_h$, as:
\begin{equation}
\begin{aligned}
    \mathbf{F}_{out} = \mathcal{I}(\mathbf{F}_{reg}, \widehat{\mathbf{F}}_h)
\end{aligned}
\end{equation}
where $\mathcal{I}$ is the aggregation function between the feature $\mathbf{F}_{reg}$ after registration and the feature $\widehat{\mathbf{F}}_h$.

\subsubsection{Loss Function}
In order to enable the network to better regress the elevation value of the instance, we use berHuloss~\cite{zwald2012berhu} as our loss function, which is expressed as
\begin{equation}
\begin{aligned}
    \mathcal{L}(x) = \left\{\begin{matrix}
     |x| & |x| \le c\\
    \frac{x^2+c^2}{2c} & |x| > c
\end{matrix}\right.  
\end{aligned}
\end{equation}
where $c$ is the judgment threshold. Specifically, $c = 0.2\times max(|predict-gt|)$ in our experiments.

\subsection{Building Extraction Network}
Unlike regular segmentation network, we only focus on the single category (buildings) of element. In the building mask extraction process, we first perform binarization preprocessing on the labels of the dataset. Specifically, the label value of the pixel where the building is located is set to 1, and the label value of the non-building pixel is set to 0. In order to better extract building masks, we use the currently popular Deeplabv3+~\cite{chen2018encoder} as our building extraction network to extract the building mask $\mathbf{M} $, which is built on top of the backbone ResNet-101. Given an input image, the building extraction network outputs a mask $\mathbf{M} $ of the building. Specifically, the building pixel value is 1, and the background is 0. During training, the update of network parameters is driven by the cross-entropy loss function.

What's more, the canny edge detection algorithm is used to extract the outlines of buildings. The extracted outlines and building masks with fine pixel positioning give the next procedures the precise building positioning.

\subsection{Building Reconstruction}
Building 3D reconstruction based on remote sensing imagery has always been a hot research topic. However, most of the existing methods are limited to a single output of the building CityGML model, and the real 3D model of buildings has not been constructed. To address this issue, we propose a building 3D reconstruction method based on building elevation information predicted by our SFFDE, Fig.~\ref{motivation}. After the aforementioned steps, we adopt the building masks $\mathbf{M}$ to extract the elevation information of the buildings, filter out the categories (such as vegetation) that are not of interest, which is formulated as:
\begin{equation}
    \mathbf{E}_{building} = \mathbf{E} \circ \mathbf{M} 
\end{equation}
where $\mathbf{E}$ and $\mathbf{E}_{building}$ are the elevation information before and after filtering. $\circ$ is the Hadamard product operation. After obtaining the elevation information of the buildings, the two procedures of point cloud reconstruction and building reconstruction (Surface Reconstruction or CityGML Reconstruction) are performed sequentially.

\textbf{Point Cloud Reconstruction.}
Generating the point cloud data of the object is a key step to obtain the three-dimensional structure model of the object. In this process, the obtained DSM prediction results of each input image patch is first fused through patch fusion to obtain a larger area of the scene. After that, in order to smooth the gap between patches, Gaussian filtering operation is employed to the DSM predicted results of each large scene remote sensing image. Then, we perform 2D-to-3D mapping of buildings based on the elevation information of large-area building clusters to generate their 3D point cloud data. The latitude and longitude coordinates corresponding to the pixels are used as the $x,y$ coordinates of the point cloud, and the elevation information is used as the $z$ coordinates of the point cloud.

\textbf{Mesh and CityGML Model.}
Mesh refers to a polygonal grid, which is a data structure used in computer graphics for modeling various irregular objects. In the face of the polygon mesh, the triangular face is the smallest unit to be divided, so it is often referred to as the triangular face. The basic components of a mesh: vertices, edges, and faces. CityGML is a data format used to construct virtual 3D city models, and is a general data model used to express 3D city templates. CityGML can not only express the graphic appearance of the city model, but also take care of the semantic representation, such as the classification and aggregation of digital terrain models, vegetation and water systems, etc. All models can be divided into five different coherent levels of detail (LOD), with increasing level of detail to obtain more details about the geometry and themes. The five consecutive levels of detail are: LOD0, LOD1, LOD2, LOD3, and LOD4. The CityGML model reconstructed in this paper is the LOD1 model in five coherent levels of detail.

\textbf{Surface Reconstruction.}
Based on the point cloud data of the building generated in the previous steps, we first normalize the point cloud to facilitate the subsequent reconstruction process. Then, we perform Poisson~\cite{kazhdan2006poisson} reconstruction on the normalized point cloud data to obtain the mesh model of the building.

\textbf{CityGML Reconstruction.}
To verify the effectiveness of our method, we also extend the SFFDE network to the algorithm for building CityGML model reconstruction in our Building3D. After our building3D performs the 3D point cloud mapping operation, we also extract the polygonal structure of the building from the image at the same time. Combining the obtained 3D point cloud and building polygon structure, we use 3dfier~\cite{3dfier} to reconstruct the building CityGML model from single-view remote sensing images.

\section{Experiments}
In this section, we first give an introduction to the datasets used by our entire framework. Next, we introduce the specific experimental setup (including evaluation protocols and completion details) in the experiments. Then, we show the performance and visualization results of our proposed SFFDE on elevation estimation. After that, a reconstruction analysis of the entire reconstruction framework Building3D is presented. Finally, to demonstrate the effectiveness of the proposed elevation semantic globalization operation and local-to-global elevation semantic registration operation, we conduct extensive ablation experiments.

\subsection{Datasets}
\subsubsection{ISPRS Vaihingen} 
There are a total of 33 slices in the ISPRS Vaihingen~\cite{vaihingen.org} dataset and each slice has approximately 2500$\times$2500 pixels. Each remote sensing image is accompanied by orthophoto images, semantic labels and digital surface models (DSM and nDSM). The ground sampling distance of each image is 9 cm, and it has three channels of near-infrared, red and green. According to the official split, 16 slices that provided ground truth are used for the training of models, and the remaining 17 slices are used for evaluation by the challenger organizer. 

\subsubsection{2019 Data Fusion Contest} 
The 2019 Data Fusion Contest (DFC2019)~\cite{bosch2019semantic} dataset currently includes approximately 100 square kilometers of coverage for Jacksonville, Florida, and Omaha, Nebraska, USA. The ground sampling distance (GSD) is about 30 cm, and each image with semantic labels and normalized DSM is 1024$\times$1024 pixel in size. The dataset provides WorldView-3 panchromatic and 8-band visible and near-infrared (VNIR) images. DFC2019 includes 26 images collected in Jacksonville, Florida, and 43 images collected in Omaha, Nebraska, USA.

\begin{table*}[htb]
    \normalsize 
    \caption{The performance of DSM estimation on ISPRS Vaihingen dataset. $\uparrow$ means that the higher the indicator value, the better the performance, and $\downarrow$ means that the lower the indicator value, the better the performance.}\label{vaihingenPerformance}
    \centering
    \begin{tabular}{l|ccc|ccc}
    \hline
    Method & Rel$\downarrow$  & RMSE$\downarrow$/m &RMSE(log)$\downarrow$ &$\delta_1\uparrow$    & $\delta_2\uparrow$   & $\delta_3\uparrow$ \\  
    \hline
    Amirkolaee \textit{et al.}~\cite{amirkolaee2019height} & 1.163 & 2.871 & 0.334 & 0.330 & 0.572 & 0.741 \\  
    IMG2DSM~\cite{ghamisi2018img2dsm}             & - & 2.58$\pm$0.09 & - & - & - & - \\ 
    D3Net~\cite{carvalho2018regression}    & 2.016 & 2.123 & - & 0.369 & 0.533 & 0.644 \\  
    Li \textit{et al.}~\cite{li2020height} & 0.314 & 1.698 & 0.155 & 0.451 & 0.817 & 0.939 \\ 
    \hline
    \bf{SFFDE + ResNet50 (ours)}  & 0.225  & 1.145  & 0.087   & \bf{0.624} & 0.841      & 0.933   \\ 
    \bf{SFFDE + ResNet101 (ours)} & \bf{0.222} & \bf{1.133}   & \bf{0.084}  & 0.595 & \bf{0.897} & \bf{0.970}   \\ 
    \hline
    \end{tabular}
\end{table*}

\subsection{Experimental Settings}
\subsubsection{Evaluation Protocols}
We use six metrics to evaluate the DSM estimation performance of our proposed method, including mean relative error (Rel), RMSE, RMSE(log), and the ratio of pixels with predicted elevation values close to the ground truth ($\delta_1, \delta_2, \delta_3$). 
Among them, mean relative error (Rel), RMSE, RMSE(log) are expressed as:
\begin{equation}
    Rel = \frac{1}{N}\sum_{i=1}^{N}\frac{|D_i-D^*_i|}{D^*_i}
\end{equation}
\begin{equation}
    RMSE = \sqrt{\frac{1}{N}\sum^{N}_{i=1}|D_i-D_i^*|^2}
\end{equation}
\begin{equation}
    RMSE(log) = \sqrt{\frac{1}{N}\sum^{N}_{i=1}|logD_i-logD_i^*|^2}
\end{equation}
where $N$ is the number of the pixels, $D_i$ is the predicted elevation value of the $i$-th pixel, and $D^*_i$ is the ground truth of the $i$-th pixel. What's more, the $\delta_i$ is expressed as:
\begin{equation}
    \delta_i = max(\frac{h_{pred}}{h_{gt}}, \frac{h_{gt}}{h_{pred}})
\end{equation}
where $h_{gt}$ and $h_{pred}$ are the ground truth and predicted elevation value.

\subsubsection{Implementation Details}
Our framework is implemented based on PyTorch Library. The momentum SGD algorithm with the momentum value set to 0.9 is employed to optimize both the DSM estimation branch and the building extraction branch. We train our model for 80000 iterations and the initial and minimum learning rate is set to 0.005 and 0.00002, respectively. The weight decay value is set to 0.0005 for regularization. Our DSM estimation branch and building extraction branch are both executed on a single NVIDIA TITAN RTX GPU with batch size set to 4. Considering the huge size of each image in both Vaihingen and DFC2019 datasets make the images unable to directly be sent to the network due to the GPU memory limit, we employ a sliding window strategy to generate small image patches with 512$\times$512 pixels. 

When evaluating accuracy, for each dataset, we only select the checkpoint saved in the last iteration (80000 iteration) for testing. The number of training iterations is set in consideration of ensuring the model is converged and stable. The choice of checkpoint is based on the weight file saved in the last iteration.

\subsection{Performance Analysis of DSM Estimation}
\subsubsection{ISPRS Vaihingen}
We compared the DSM estimation performance by our SFFDE with other prediction networks, such as Amirkolaee \textit{et al.}~\cite{amirkolaee2019height}, IMG2DSM~\cite{ghamisi2018img2dsm}, D3Net~\cite{carvalho2018regression}, and Li \textit{et al.}~\cite{li2020height}. In Table~\ref{vaihingenPerformance}, the comparisons between our SFFDE network and these methods on the ISPRS Vaihingen dataset are given. It is clear that our SFFDE network achieves the highest performance for the elevation estimation task. Regardless of choosing ResNet50 or ResNet101 as the backbone, our SFFDE outperforms state-of-the-art methods on all metrics. Rel, RMSE and RMSE(log) are indicators that describe the prediction error of the model. Among these metrics, with the ResNet101 backbone, our SFFDE achieves $0.222$ Rel, $1.133$ RMSE and $0.084$ RMSE(log), achieving the best performance. This shows that the prediction results of our SFFDE maintain a high consistency with the ground truth. What's more, $\delta_1, \delta_2$ and $\delta_3$ are indicators describing the prediction accuracy of the model. Note that for the indicators ($\delta_1, \delta_2, \delta_3$), SFFDE achieves the highest $\delta_i$ value, which strongly demonstrates the elevation value predicted by our SFFDE is close to the ground truth elevation value. This again proves that the local and global semantic registration implemented by SFFDE can improve the accuracy of elevation prediction.

The reason is that our SFFDE can use elevation semantic globalization (ESG) to achieve global feature extraction and local-to-global elevation semantic registration (L2G-ESR) to achieve global and local feature registration. Based on the concept of elevation semantic flow, we can well explicitly model the semantic change rate of semantic features describing elevation across different resolutions.

\subsubsection{DFC2019}
In Table~\ref{DFC2019Performance}, our SFFDE network is compared with other state-of-the-art DSM estimation methods (including D3Net~\cite{carvalho2018regression}, DORN~\cite{fu2018deep}, and FastDepth~\cite{wofk2019fastdepth}) on DFC2019 dataset. Clearly, our SFFDE outperforms all existing elevation estimation methods on the DFC2019 dataset. Table~\ref{DFC2019Performance} displays the evaluation estimation results of our SFFDE network. Whether we choose ResNet50 or ResNet101 as our backbone, SFFDE again outperforms state-of-the-art methods on all indicators. For the ResNet101 backbone, SFFDE improves FastDepth by $\bf{0.108}$ ($0.492$ vs. $0.384$) on $\delta_1$, $\bf{0.081}$ ($0.782$ vs. $0.701$) on $\delta_2$, and $\bf{0.033}$ ($0.908$ vs. $0.875$) on $\delta_3$, which are large margins for the challenging DSM estimation problem. Compared with the ISPRS Vaihingen dataset, the DFC2019 dataset has more complex scenes and diverse instances, which undoubtedly brings great difficulties to the depth estimation problem. However, our SFFDE not only performs the best on the ISPRS Vaihingen dataset, but also has the highest performance on the DFC2019 dataset, which once again proves that our proposed feature registration based on the elevation semantic flow field can well improve complex scenes elevation prediction performance.

\begin{table*}[htb]
    \normalsize 
    \caption{The performance of DSM estimation on DFC2019 dataset. $\uparrow$ means that the higher the indicator value, the better the performance, and $\downarrow$ means that the lower the indicator value, the better the performance.}\label{DFC2019Performance}
    \centering
    \begin{tabular}{l|cc|ccc}
    \hline
    Method   & Rel$\downarrow$  &RMSE(log)$\downarrow$ &$\delta_1\uparrow $ & $\delta_2\uparrow$ & $\delta_3\uparrow$ \\  
    \hline
    D3Net~\cite{carvalho2018regression}     & 0.526        & 0.208  & 0.256  & 0.635 & 0.846   \\  
    DORN~\cite{fu2018deep}                  & 0.488        & 0.200  & 0.317  & 0.646 & 0.859   \\ 
    FastDepth~\cite{wofk2019fastdepth}      & 0.383        & 0.189  & 0.384  & 0.701 & 0.875   \\ 
    \hline
    \bf{SFFDE + ResNet50 (ours)}       & \bf{0.272}  & 0.029  & \bf{0.601} & 0.778  & 0.882   \\ 
    \bf{SFFDE + ResNet101 (ours)}      & 0.330      & \bf{0.024}  & 0.492  & \bf{0.782}  & \bf{0.908}  \\ 
    \hline
    \end{tabular}
\end{table*}

\begin{figure*}
    \begin{center}
    \includegraphics[width=1.0\linewidth]{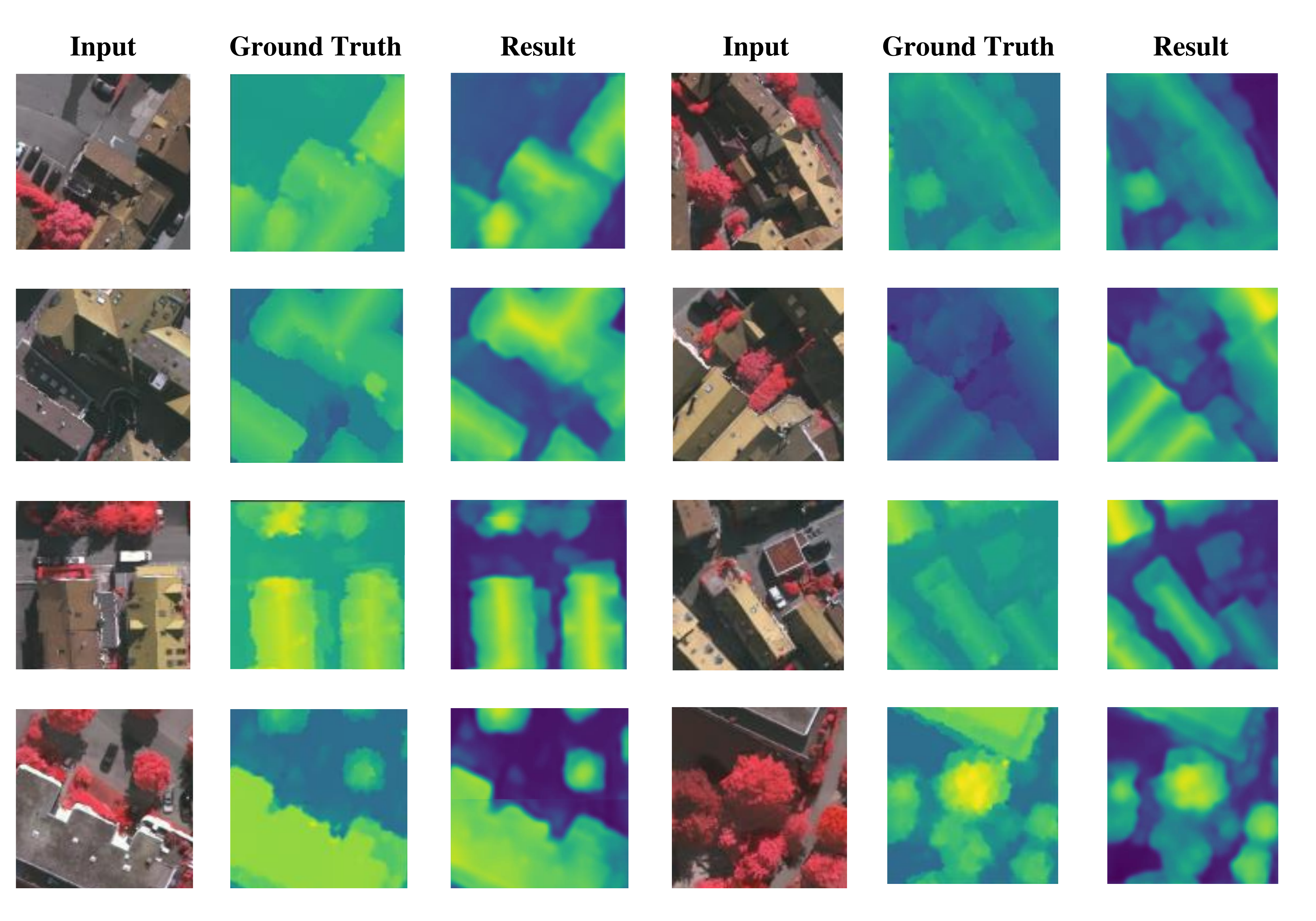}
    \end{center}
    \caption{Visualizations of the predicted elevation results (512 $\times$ 512 patches) of our SFFDE network on ISPRS Vaihingen dataset.}
    \label{DSMestimation_Patch}
\end{figure*}
\begin{figure*}
    \begin{center}
    \includegraphics[width=1.0\linewidth]{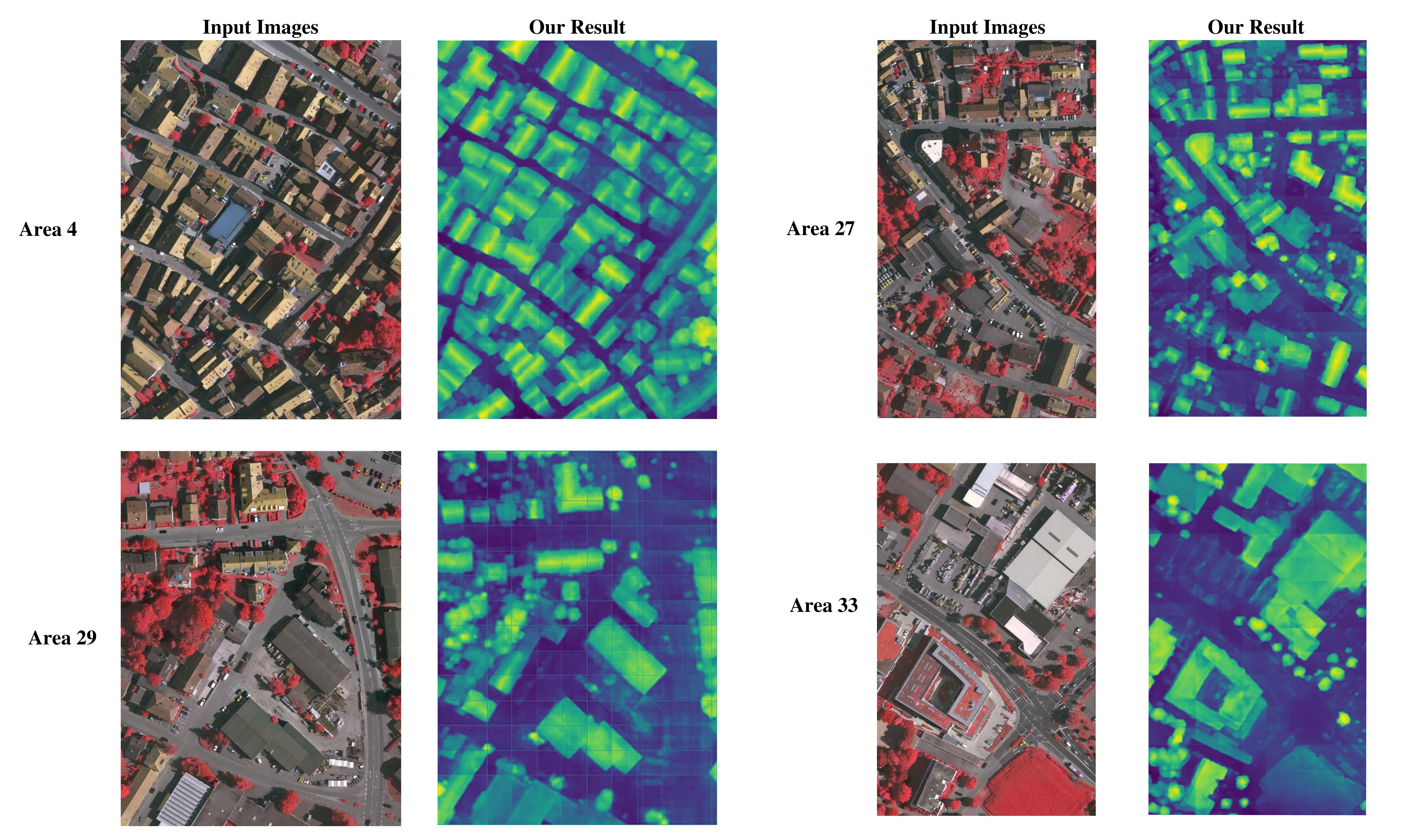}
    \end{center}
    \caption{Visualizations of the predicted elevation results (large areas) of our SFFDE network on ISPRS Vaihingen dataset.}
    \label{DSMestimation}
\end{figure*}
\subsection{Visualization Analysis}
\subsubsection{DSM Visualizations on ISPRS Vaihingen}
As shown in Fig. ~\ref{DSMestimation_Patch}, local area visualizations of the DSM estimation results of SFFDE is given on the ISPRS Vaihingen dataset. From the visualized results, we can conclude that our SFFDE has high accuracy for building elevation prediction, and the prediction results are basically consistent with the ground truth. Furthermore, the visualization clearly shows that our predicted instance elevations are well-defined. At instance boundaries, our method has small prediction errors and aliasing.

As shown in Fig. ~\ref{DSMestimation}, we also present the visualization results of the elevation prediction for a large area of the ISPRS Vaihingen dataset. The images of three large scenes (Area 4 contains a large number of buildings, Areas 27 and 29 have both regular and irregular texture instances, and Area 33 has more interior details of buildings) are selected as the inputs of our SFFDE and the prediction results are visualized. From the visualization results of Area 4, our method is able to make good elevation predictions for areas with dense buildings. In these densely built areas, the buildings are all structures with high roofs in the center and low on both sides. Nonetheless, our visualizations fit this phenomenon well, yielding high predictive performance. This is highlighted in the middle of the roof in the picture, and the sides are darker to demonstrate that. Areas 27 and 29 contain a large number of regular and irregular instances such as buildings and trees. For the regions where these regular and irregular instances coexist, this requires the model to have high generalization performance. From the visualization results, our SFFDE achieves superior elevation prediction performance and can guarantee high accuracy for both regular and irregular texture instances. This benefits from the higher performance of our proposed elevation semantic flow-based feature registration. Furthermore, our SFFDE can predict the detailed structure of buildings well, as can be seen from the results for Area 33. On top of complex buildings, smooth roofs and complex structures coexist. In our prediction results, the high accuracy of the smooth roof prediction is reflected in the smoothness of the elevation prediction of the roof, which is proved by the basically consistent colors of the smooth areas in the figure. In addition, in the region of complex structure, our SFFDE gives fine boundary structure prediction results. This is superior to other advanced methods.

\begin{figure*}
    \begin{center}
    \includegraphics[width=1.0\linewidth]{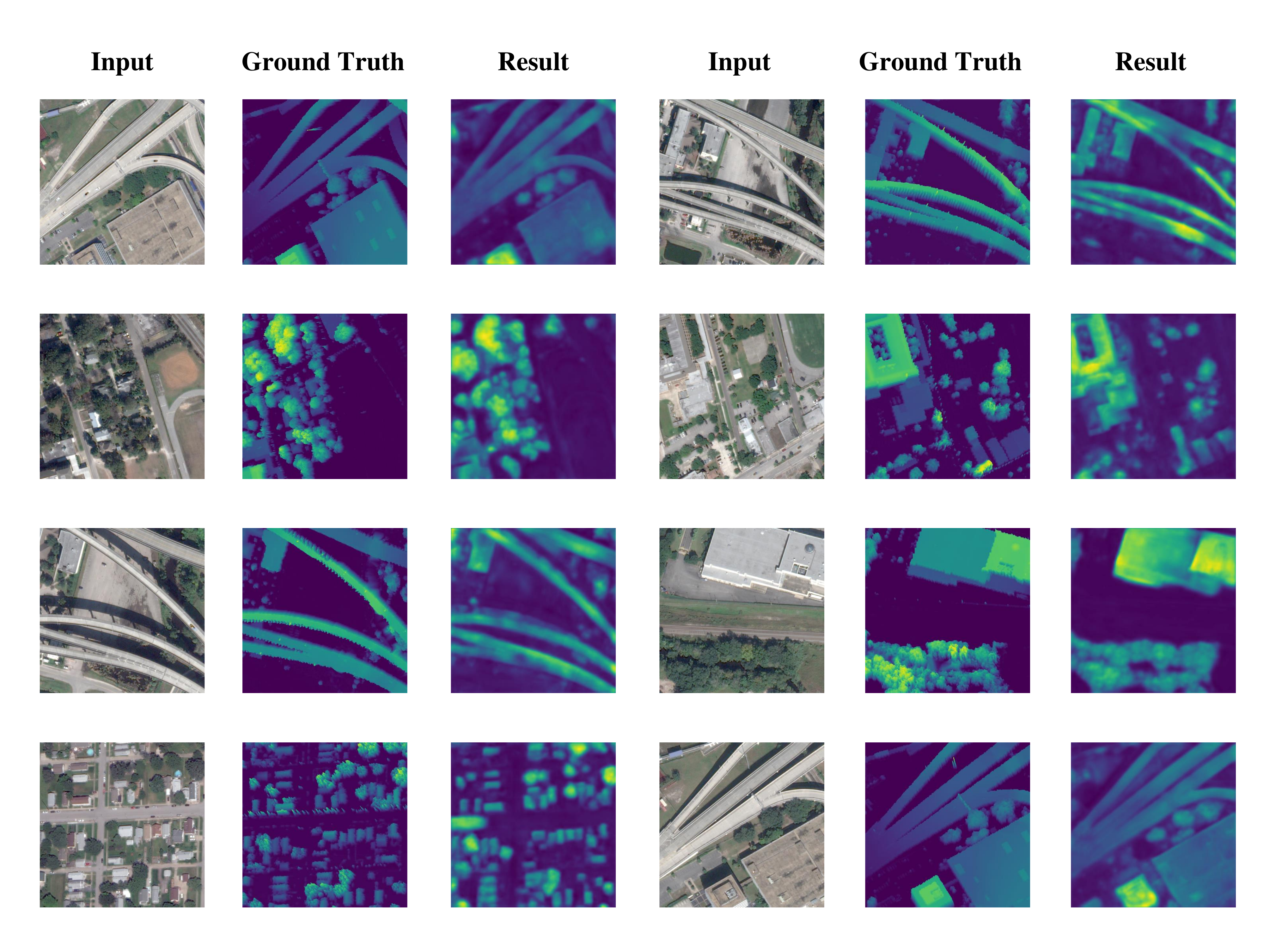}
    \end{center}
    \caption{Visualizations of the predicted elevation results (512 $\times$ 512 patches) of our SFFDE network on DFC2019 dataset.}
    \label{DSMEstimation_patch_us3d}
\end{figure*}
\begin{figure*}
    \begin{center}
    \includegraphics[width=1.0\linewidth]{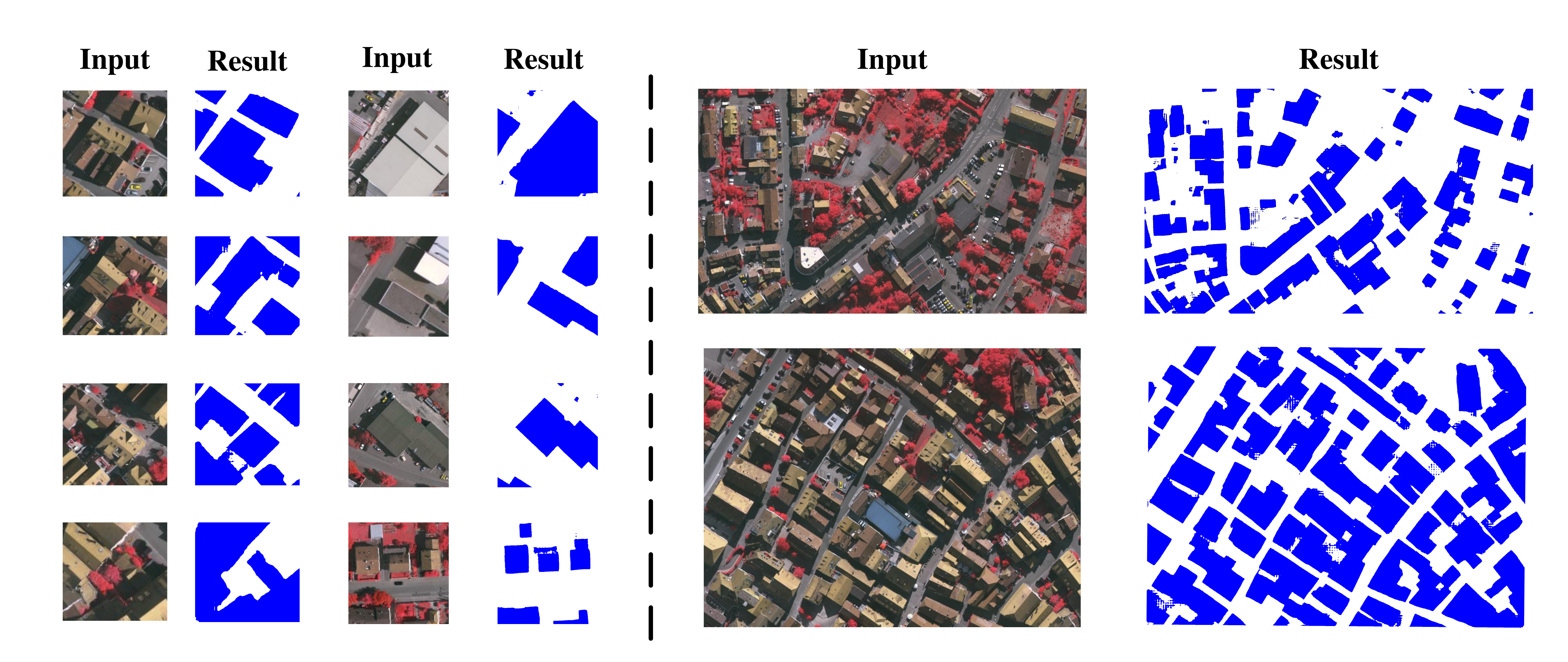}
    \end{center}
    \caption{Visualizations of the building extraction network in our Building3D framework. The left part is the building extraction results of the image patches, and the right part is the extraction results of large area buildings.}
    \label{BuildingExtraction}
\end{figure*}

\subsubsection{DSM Visualizations on DFC2019}
To verify the performance of our SFFDE, we also conduct experiments on the DFC2019 dataset and give high performance. Furthermore, we present the prediction visualization results of our SFFDE on the DFC2019 dataset in Fig. ~\ref{DSMEstimation_patch_us3d}. Since DFC2019 contains complex instances such as bridges, buildings, trees, etc., our visualization results all contain these complex instances. It can be seen from the visualization results that the elevation information of various instances we predicted is basically consistent with the ground truth. Furthermore, for regularly textured objects such as buildings and bridges, our prediction results have clear boundary information and smooth internal structures, which are extremely challenging problems in elevation prediction tasks. Nonetheless, our SFFDE exhibits high prediction performance, which validates the effectiveness of our method. In addition, we also achieved high prediction accuracy for objects with irregular textures such as trees.
\begin{figure*}
    \begin{center}
    \includegraphics[width=1.0\linewidth]{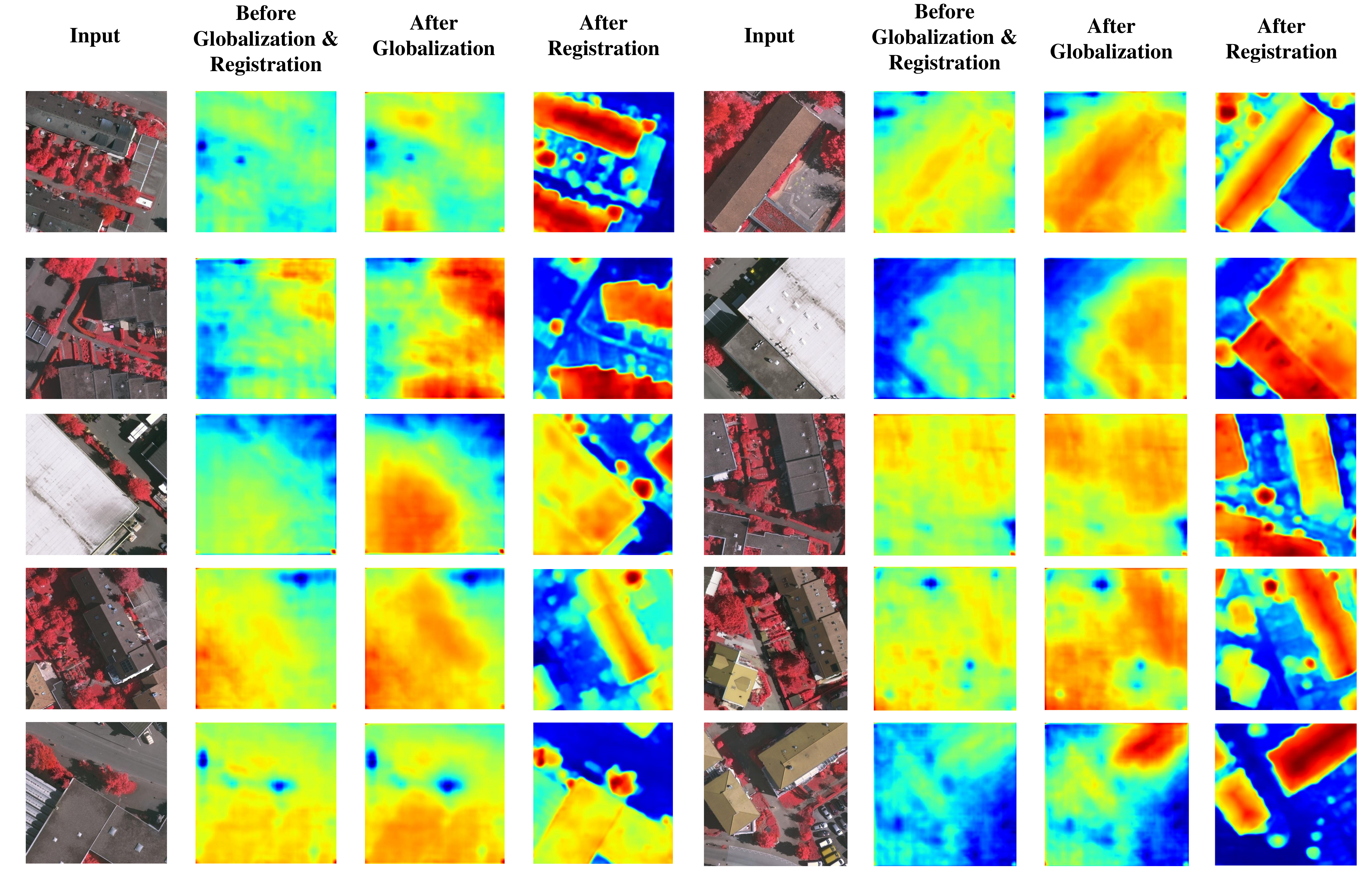}
    \end{center}
    \caption{Visualizations of the feature maps after globalization and registration. The first column is the input of our SFFDE, the second column is the feature before globalization and registration, the third column is the feature map after globalization, and the fourth column is the feature map after registration.}
    \label{FeatureMap}
\end{figure*}

\begin{figure*}
    \begin{center}
    \includegraphics[width=1.0\linewidth]{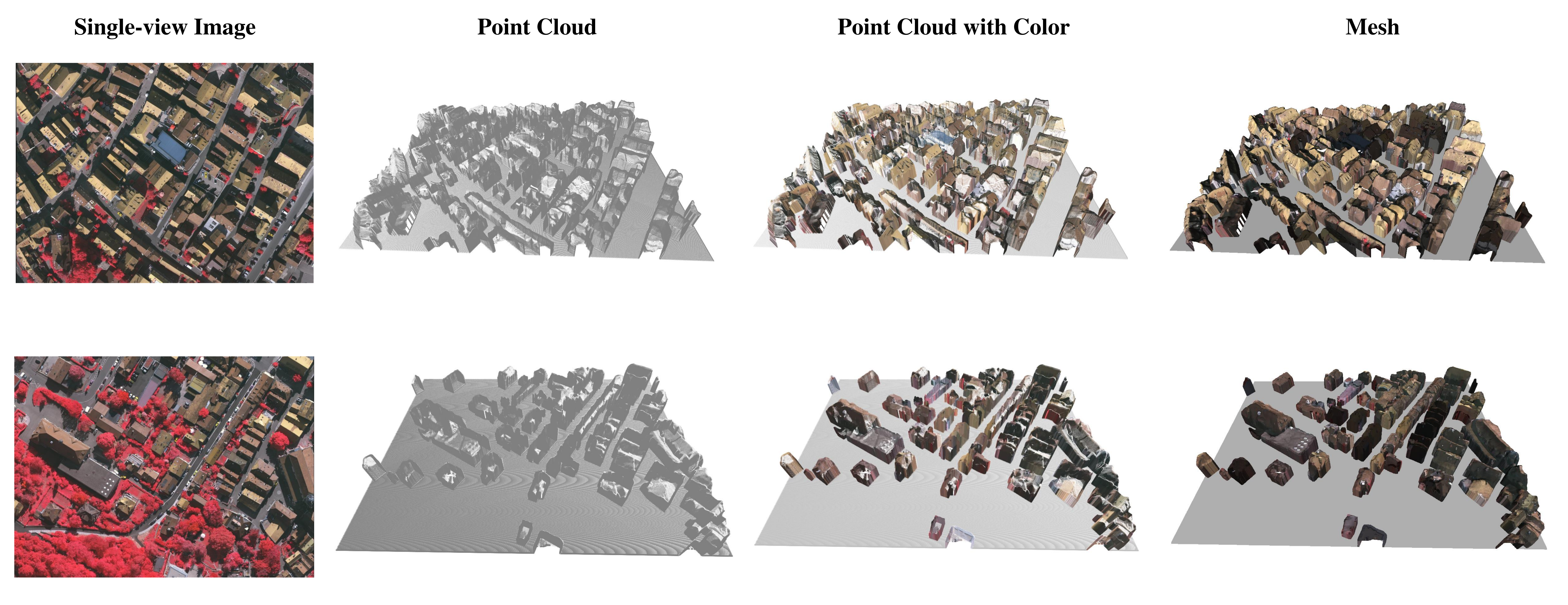}
    \end{center}
    \caption{Visualizations of point cloud and surface mesh of the input single-view image. The first column is the input single-view images, the second column is the reconstructed point clouds, the third column is the reconstructed point clouds with color, and the last column is the reconstructed mesh models of the input images.}
    \label{PC_MeshVisual}
\end{figure*}

\subsubsection{Feature Map Visualizations}
Feature map visualizations on the ISPRS Vaihingen dataset are given as shown in Fig. ~\ref{FeatureMap}. As noted in the figure, the visualization consists of four columns, which are the input image, the feature map before globalization and registration, the feature map after globalization, and the feature map after global-local semantic registration. From the feature map after globalization, we can infer that the feature obtains a global receptive field, not only limited to local perception. In addition, the registered feature map can clearly observe the object and its detailed information, and the internal features of the instance are smooth and the pixel values tend to be continuous. This matches well with better regression problems. Therefore, from the visualization results of the two-part feature maps, it can be seen that our network can simultaneously ensure the feature's ability to perceive the global and the local perception of details, and achieve a better balance between global and local features.

\begin{table*}[htb]
    \normalsize            
    \caption{Ablation studies on ISPRS Vaihingen dataset. `ESG' denotes Elevation Semantic Globalization. `L2G-ESR' represents Local-to-Global Elevation Semantic Registration. $\uparrow$ means that the higher the indicator value, the better the performance, and $\downarrow$ means that the lower the indicator value, the better the performance.}\label{Ablation}
    \centering
    \begin{tabular}{l|ccc|ccc}
    \hline
    Method  & Rel$\downarrow$ & RMSE$\downarrow$/m &RMSE(log)$\downarrow$ &$\delta_1\uparrow$ & $\delta_2\uparrow$ & $\delta_3\uparrow$ \\  
    \hline
    \multicolumn{7}{c}{ResNet-50}\\
    \hline
    Baseline                     & 0.367  & 1.330  & 0.135   & 0.357  & 0.618  & 0.819\\    
    +ESG                         & 0.277  & 1.188  & 0.109   & 0.542  & 0.800  & 0.915 \\    
    +ESG + L2G-ESR (ours)        & \bf{0.225} & \bf{1.145} & \bf{0.087}  & \bf{0.624} & \bf{0.841} & \bf{0.933}\\    
    \hline
    \multicolumn{7}{c}{ResNet-101}\\
    \hline
    Baseline                     & 0.358  & 1.293  & 0.130   & 0.374      & 0.701      & 0.870  \\
    +ESG                         & 0.276  & 1.282  & 0.111   & 0.534      & 0.843      & 0.952  \\
    +ESG + L2G-ESR (ours)  & \bf{0.222}     & \bf{1.133}  & \bf{0.084}  & \bf{0.595} & \bf{0.897} & \bf{0.970} \\    
    \hline
    \end{tabular}
\end{table*}

\subsection{Building3D Reconstruction Analysis}
\subsubsection{Building Extraction}
As shown in Fig. ~\ref{BuildingExtraction}, the visualization results of the building extraction network in Building3D framework are given. We not only visualize the local building area in Vaihingen, but also visualize the extraction results of building groups in a large area. Based on the superior building extraction network, we can see that the extracted building area is basically consistent with the actual building area, achieving extremely high building extraction performance. This lays a good foundation for subsequent building elevation extraction, building point cloud reconstruction and surface reconstruction.

\subsubsection{Surface Reconstruction}
Given a single-view remote sensing image of an area, after the building elevation prediction and building extraction steps mentioned above are completed, the next process is to reconstruct the building in 3D. Based on the extracted building areas, we filter the predicted elevation information to obtain building elevations. The visualization results of building surface reconstruction based on single-view remote sensing images are given in the Fig. ~\ref{PC_MeshVisual}. Given a single-view remote sensing image, our Building3D outputs a point cloud and mesh of the building. From Fig. ~\ref{PC_MeshVisual} we can see that the building has been converted into a 3D structure, while other elements such as cars, vegetation, etc. have not been converted. The first column in Fig. ~\ref{PC_MeshVisual} shows the input single-view image. On the basis of the obtained building elevations, we perform 3D mapping on the recovered elevation information to obtain 3D point clouds. The results of the 3D point cloud visualization of the mapped buildings are given in the second column of the figure. In order to better show the point cloud results, we also attached the color to the point cloud, which is given in the third column of the figure. Based on the recovered 3D point cloud, the surface reconstruction results of the building are shown in the fourth column of the figure. Clearly, our framework generates high-quality 3D point cloud data and builds a 3D model when there is only the single-view image. This provides an insightful idea for rapid 3D reconstruction of large-scale buildings.
\begin{figure}
    \begin{center}
    \includegraphics[width=1.0\linewidth]{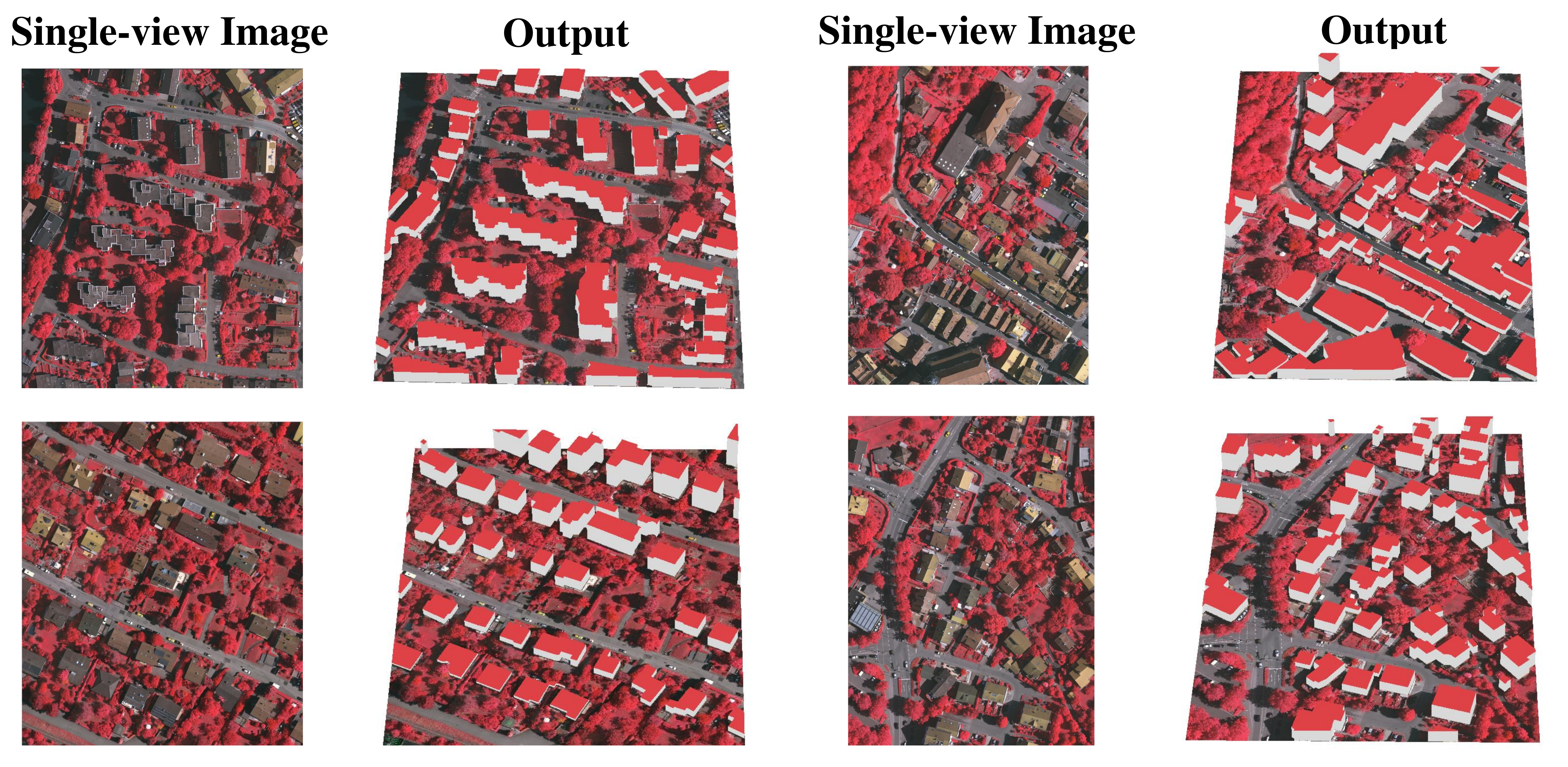}
    \end{center}
    \caption{Visualizations of LOD1 model of buildings in the input single-view image.}
    \label{CityGML}
\end{figure}

\begin{figure*}
    \begin{center}
    \includegraphics[width=0.95\linewidth]{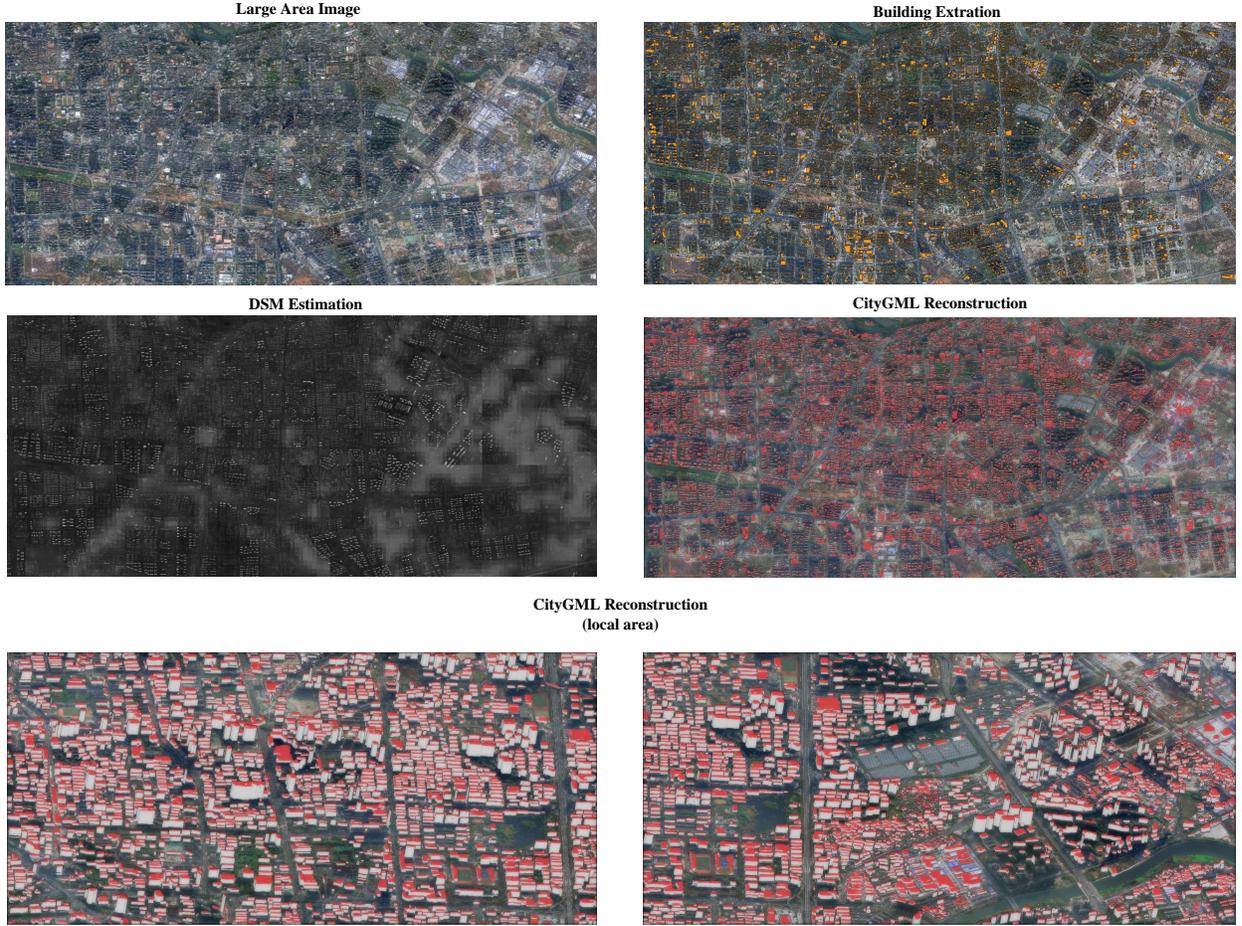}
    \end{center}
    \caption{Visualization Results of large area building reconstruction. The large image is obtained from the urban area of Hefei City, Anhui Province, China.}
    \label{extension}
\end{figure*}

\subsubsection{CityGML Reconstruction}
For the reconstruction of the CityGML model of the buildings, our Building3D adopts the 3dfier~\cite{3dfier} approach. The input is the point cloud data obtained in the previous steps and the polygon structure of the buildings, and the output is the LOD1 model of the buildings. The reconstruction results are given in Fig. ~\ref{CityGML}. Obviously, the building elevation information predicted by our SFFDE can restore the building elevation well, laying a foundation for the 3D reconstruction of different forms of buildings.
\begin{figure}
    \begin{center}
    \includegraphics[width=0.7\linewidth]{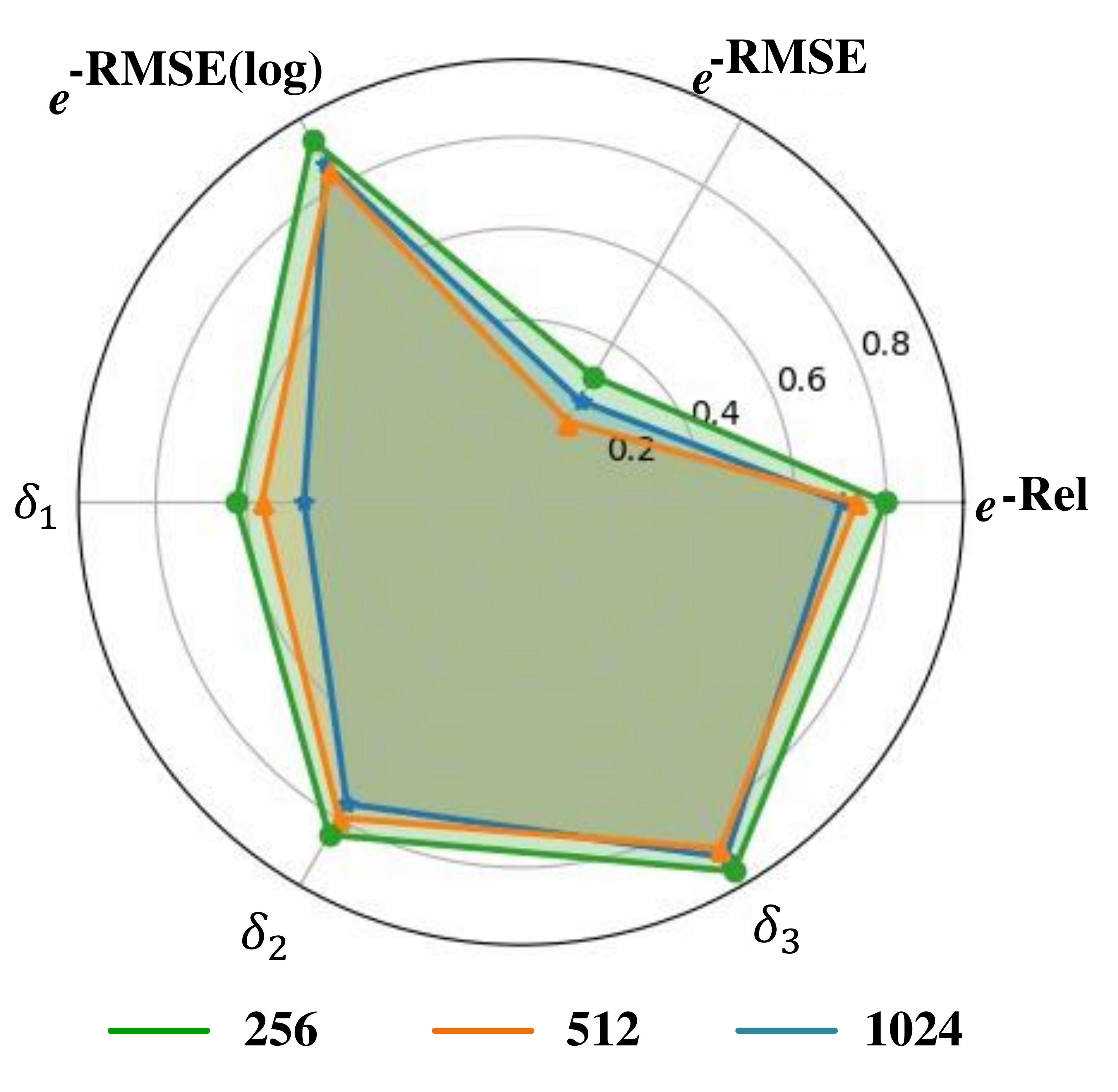}
    \end{center}
    \caption{Ablation studies of the channel number of elevation semantic globalization (ESG).}
    \label{AblationChannel}
\end{figure}

\subsection{Extension}
\subsubsection{Large area building reconstruction}
We test the robustness experiments on remote sensing images of the whole urban area in Hefei, Anhui Province, China. As shown in Fig. ~\ref{extension}, we give the remote sensing images of the entire area, the building extraction results, the DSM prediction results, and the building LOD1 model reconstruction results, respectively.

\subsection{Ablation Study}
We conduct ablation studies on ISPRS Vaihingen dataset to verify the effectiveness of our SFFDE. PSPNet~\cite{zhao2017pyramid} is selected as the baseline.

\subsubsection{ESG}
As shown in Table~\ref{Ablation}, with elevation semantic globalization, our SFFDE with ResNet101 as the backbone improves the DSM estimation performance by $\bf{0.082}$ ($0.276$ vs. $0.358$) on Rel, $\bf{0.011}$ ($1.282$ vs. $1.293$) on RMSE, and $\bf{0.019}$ ($0.111$ vs. $0.130$) on RMSE(log), which validates that the features with global dependency improve estimation performance more significantly. Before globalization, features of the network are limited to local perception, so that the perception of texture regular and irregular instances cannot be provided at the same time. However, after adding ESG, the perception ability of the network is extended from local to global, laying the foundation for subsequent high-level semantic extraction. At the same time, the introduction of ESG also provides a priori global features for the subsequent registration of global and local features.
\begin{figure*}
    \begin{center}
    \includegraphics[width=1.0\linewidth]{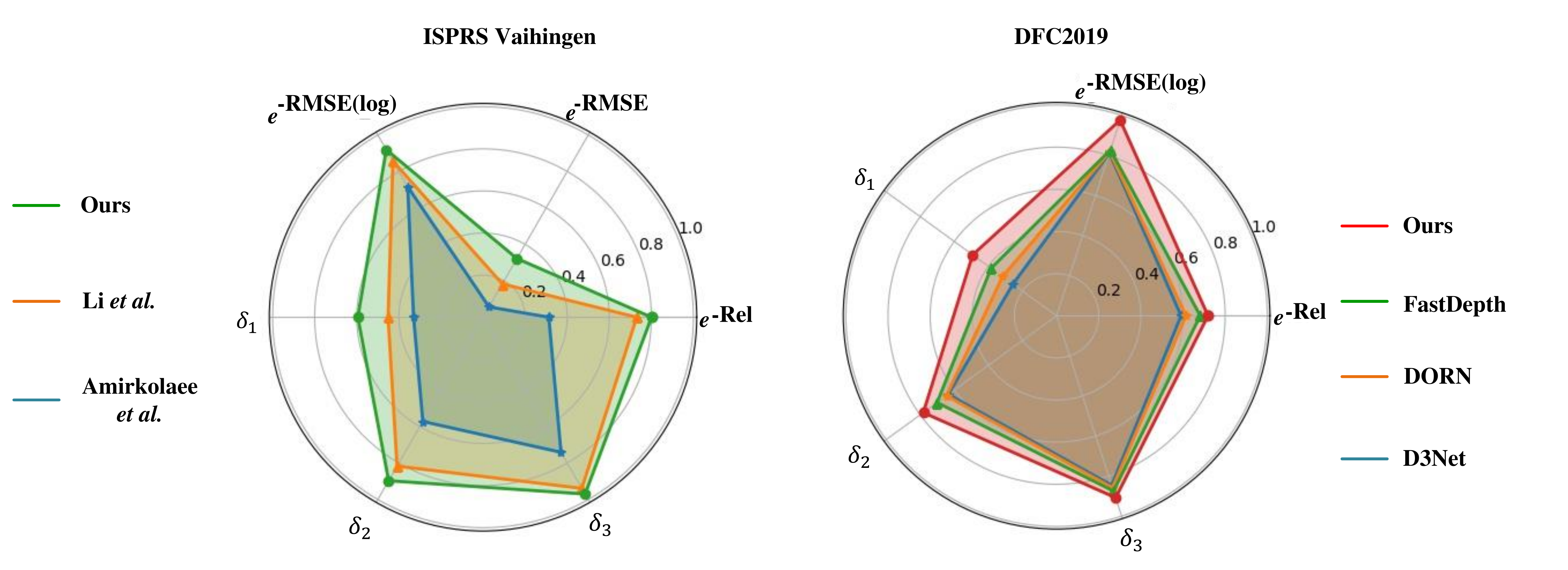}
    \end{center}
    \caption{Performance spider plots on ISPRS Vaihingen dataset and DFC2019 dataset.}
    \label{PerformanceSpider}
\end{figure*}

\textbf{Channel Number of ESG.}
In order to better utilize the globalization ability of ESG operations, we conduct ablation experiments on the number of input feature channels. For better visualization, we show spider plots with different channel numbers, as shown in Fig. ~\ref{AblationChannel}. In order to better display the spider chart, we have performed negative index operations ($e^{-x}$, $x$ is the corresponding indicator) on the three indicators of Rel, RMSE, and RMSE(log). Based on this, the six indicators are all closer to the periphery on behalf of higher performance. As can be seen from the Fig. ~\ref{AblationChannel}, when the number of channels is 256, the values of all indicators are located at the outermost periphery, which means that ESG has the best performance at this time.

\subsubsection{Backbone Selection} 
To verify the influence of different backbones on our method, we conduct experiments on resnet50 and resnet101 respectively. As shown in the table, we add the proposed elevation semantic globalization operation and L2G-ESR operation on the basis of resnet50 and resnet101, respectively. We make a horizontal comparison of the backbone, then we come to the conclusion that the prediction accuracy of selecting resnet101 as the backbone as a whole is higher than resnet50. Similarly, after adding ESG and L2G-ESR, the overall performance of the resnet101-based network is also better than that of resnet50. This is consistent with our prior knowledge perception.
\begin{table*}[htb]
    \normalsize            
    \caption{Ablation studies on aggregation operation. `Concat joint Conv' denotes concatenation joint convolution. `Element-wise Add' represents the element-wise addition operation. $\uparrow$ means that the higher the indicator value, the better the performance, and $\downarrow$ means that the lower the indicator value, the better the performance.}\label{Aggregation}
    \centering
    \begin{tabular}{l|ccc|ccc}
    \hline
    Aggregation Operation  & Rel$\downarrow$ & RMSE$\downarrow$/m &RMSE(log)$\downarrow$ &$\delta_1\uparrow$ & $\delta_2\uparrow$ & $\delta_3\uparrow$ \\  
    \hline
    \hline
    Concat joint Conv   & 0.263  & 1.252  & 0.101   & 0.563  & 0.718  & 0.919\\    
    Element-wise Add & \bf{0.225} & \bf{1.145} & \bf{0.087}  & \bf{0.624} & \bf{0.841} & \bf{0.933}\\    
    \hline        
    \end{tabular}
\end{table*}

\begin{table*}[htb]
    \normalsize            
    \caption{Ablation studies on loss functions. $\uparrow$ means that the higher the indicator value, the better the performance, and $\downarrow$ means that the lower the indicator value, the better the performance.}\label{lossfunction}
    \centering
    \begin{tabular}{l|ccc|ccc}
    \hline
    Loss Function  & Rel$\downarrow$ & RMSE$\downarrow$/m &RMSE(log)$\downarrow$ &$\delta_1\uparrow$ & $\delta_2\uparrow$ & $\delta_3\uparrow$ \\  
    \hline
    \hline
    L1Loss    & 0.298  & 1.265  & 0.100   & 0.548  & 0.820  & 0.912\\    
    MSELoss   & 0.293  & 1.210  & 0.099   & 0.543  & 0.836  & \bf{0.933}\\    
    berHuLoss & \bf{0.225} & \bf{1.145} & \bf{0.087}  & \bf{0.624} & \bf{0.841} & \bf{0.933}\\    
    \hline
    \end{tabular}
\end{table*}

\subsubsection{L2G-ESR} 
In Table~\ref{Ablation}, L2G-ESR further improves the performance by $\bf{0.054}$ ($0.222$ vs. $0.276$) on Rel, $\bf{0.149}$ ($1.133$ vs. $1.282$) on RMSE, and $\bf{0.027}$ ($0.084$ vs. $0.111$) on RMSE(log), which validates that L2G-ESR implements the registration and trade-off of local and global features. This clearly demonstrates the superiority of SFFDE over other methods on local and global feature representation. After L2G-ESR, the features of the network are locally and globally registered through the concept of a defined elevation semantic flow. This enables the network to perceive the surrounding pixel features finely in both global and local structure.

\textbf{Aggregation Operation $\mathcal{I}$.}
As shown in Eq. 13, there are two options in our L2G-ESR feature aggregation: (1) concatenate operation combined with convolution; (2) element-wise addition. We conduct ablation experiments for these two operations, and the experimental results are shown in Table ~\ref{Aggregation}. As can be seen from the table, element-wise addition achieves higher performance. Compared with the operation of concatenation joint convolution, element-by-element addition not only does not add redundant parameters, but also enables the network to run efficiently.

\subsubsection{Loss Function} 
In order to choose the best loss function as the learning direction of the network, we choose L1loss, MSEloss, and berHuLoss for ablation experiments. As shown in Table ~\ref{lossfunction}, choosing berHuloss as the loss function for network training enables the network to achieve the best prediction performance. Nonetheless, choosing L1loss and MSEloss also achieves impressive performance, in contrast to the baseline. This side reflects the effectiveness of our ESG and L2G-ESR.

\subsubsection{Performance Spider Plots} 
To illustrate the predicted performance of our SFFDE, we visualize the spider plots of the performance, Fig. ~\ref{PerformanceSpider}. We compare with Li \textit{et al.}~\cite{li2020height} and Amirkolaee \textit{et al.}~\cite{amirkolaee2019height} on the ISPRS dataset. On the DFC2019 dataset, comparisons with D3Net~\cite{carvalho2018regression}, DORN~\cite{fu2018deep} and FastDepth~\cite{wofk2019fastdepth} are given. Since the three indicators of Rel, RMSE, and RMSE(log) are lower, the better, so we carry out negative index processing ($e^{-x}$). Based on this, these six indicators are all as high as possible. We present the performance spider plots for the ISPRS Vaihingen dataset and the DFC2019 dataset, respectively. Since the higher the indicator value, the better the performance, the closer the indicator value of the spider plot is to the periphery, the higher the performance. It is obvious that our SFFDE is at the outermost periphery of the spider plot, whether it is the ISPRS Vaihingen dataset or the DFC2019 dataset. Therefore, this shows that our SFFDE outperforms existing methods in either metric.

\section{Conclusion}
We propose a framework (Building3D) for creating 3D building models from single-view remote sensing imagery. Our Building3D is rooted in the proposed SFFDE network to achieve globalization of semantics and registration of global features with local features through the proposed ESG and L2G-ESR. Extensive experiments on the commonly used ISPRS Vaihingen and DFC2019 datasets demonstrate the superiority of SFFDE for DSM estimation, providing accurate elevation information for building reconstruction and $\delta_1$, $\delta_2$ and $\delta_3$ metrics of our SFFDE are improved to 0.595, 0.897 and 0.970. Building3D achieves 3D reconstruction of large-area buildings in a single-view image by utilizing SFFDE elevation estimation, building mask extraction, point cloud reconstruction and building reconstruction, which is in sharp contrast to other methods based on multi-view images. Building3D not only reduces data acquisition costs, but also enables rapid and large-area building reconstruction. As a novel-and-efficient method, Building3D provides a fresh perspective into challenging 3D reconstruction of buildings. Although the 3D reconstruction effect is still greatly improved, we will continue to study to generate more refined 3D reconstruction results.

For the lattice-like stripes phenomenon, we will work in future research. First, we will focus on large-scale images as input to build a neural network for learning, which greatly reduces the number of patch edges. In addition, we will also work on the design of more accurate inter-patch fusion methods. For the overall network architecture, although our Building3D adopts the method of elevation estimation and building extraction parallel reasoning in the overall architecture to improve efficiency, the number of network parameters has increased. In view of this, we plan to introduce a multi-task learning method in the follow-up research, by inputting a single image block and simultaneously performing multi-task output (elevation regression head and building extraction head).

\bibliographystyle{IEEEtran}
\bibliography{egbib}

\begin{thebibliography}{10}
\providecommand{\url}[1]{#1}
\csname url@samestyle\endcsname
\providecommand{\newblock}{\relax}
\providecommand{\bibinfo}[2]{#2}
\providecommand{\BIBentrySTDinterwordspacing}{\spaceskip=0pt\relax}
\providecommand{\BIBentryALTinterwordstretchfactor}{4}
\providecommand{\BIBentryALTinterwordspacing}{\spaceskip=\fontdimen2\font plus
\BIBentryALTinterwordstretchfactor\fontdimen3\font minus
  \fontdimen4\font\relax}
\providecommand{\BIBforeignlanguage}[2]{{%
\expandafter\ifx\csname l@#1\endcsname\relax
\typeout{** WARNING: IEEEtran.bst: No hyphenation pattern has been}%
\typeout{** loaded for the language `#1'. Using the pattern for}%
\typeout{** the default language instead.}%
\else
\language=\csname l@#1\endcsname
\fi
#2}}
\providecommand{\BIBdecl}{\relax}
\BIBdecl

\bibitem{li2020geometry}
X.~Li, L.~Wang, and Y.~Fang, ``Geometry-aware segmentation of remote sensing
  images via implicit height estimation,'' \emph{arXiv preprint
  arXiv:2006.05848}, 2020.

\bibitem{mahdi2020aerial}
E.~Mahdi, Z.~Ziming, and H.~Xinming, ``Aerial height prediction and refinement
  neural networks with semantic and geometric guidance,'' \emph{arXiv preprint
  arXiv:2011.10697}, 2020.

\bibitem{mou2018im2height}
L.~Mou and X.~X. Zhu, ``Im2height: Height estimation from single monocular
  imagery via fully residual convolutional-deconvolutional network,''
  \emph{arXiv preprint arXiv:1802.10249}, 2018.

\bibitem{wang2020boundary}
Y.~Wang, W.~Ding, R.~Zhang, and H.~Li, ``Boundary-aware multitask learning for
  remote sensing imagery,'' \emph{IEEE Journal of selected topics in applied
  earth observations and remote sensing}, vol.~14, pp. 951--963, 2020.

\bibitem{batra2012learning}
D.~Batra and A.~Saxena, ``Learning the right model: Efficient max-margin
  learning in laplacian crfs,'' in \emph{2012 IEEE Conference on Computer
  Vision and Pattern Recognition}.\hskip 1em plus 0.5em minus 0.4em\relax IEEE,
  2012, pp. 2136--2143.

\bibitem{saxena2005learning}
A.~Saxena, S.~Chung, and A.~Ng, ``Learning depth from single monocular
  images,'' \emph{Advances in neural information processing systems}, vol.~18,
  2005.

\bibitem{hu2022pseudo}
W.-S. Hu, H.-C. Li, R.~Wang, F.~Gao, Q.~Du, and A.~Plaza, ``Pseudo
  complex-valued deformable convlstm neural network with mutual attention
  learning for hyperspectral image classification,'' \emph{IEEE Transactions on
  Geoscience and Remote Sensing}, vol.~60, pp. 1--17, 2022.

\bibitem{li2022a3clnn}
H.-C. Li, W.-S. Hu, W.~Li, J.~Li, Q.~Du, and A.~Plaza, ``A3clnn: Spatial,
  spectral and multiscale attention convlstm neural network for multisource
  remote sensing data classification,'' \emph{arXiv preprint arXiv:2204.04462},
  2022.

\bibitem{lucas1981iterative}
B.~D. Lucas, T.~Kanade \emph{et~al.}, \emph{An iterative image registration
  technique with an application to stereo vision}.\hskip 1em plus 0.5em minus
  0.4em\relax Vancouver, 1981, vol.~81.

\bibitem{saxena20083}
A.~Saxena, S.~H. Chung, and A.~Y. Ng, ``3-d depth reconstruction from a single
  still image,'' \emph{International journal of computer vision}, vol.~76,
  no.~1, pp. 53--69, 2008.

\bibitem{amirkolaee2019height}
H.~A. Amirkolaee and H.~Arefi, ``Height estimation from single aerial images
  using a deep convolutional encoder-decoder network,'' \emph{ISPRS journal of
  photogrammetry and remote sensing}, vol. 149, pp. 50--66, 2019.

\bibitem{wofk2019fastdepth}
D.~Wofk, F.~Ma, T.-J. Yang, S.~Karaman, and V.~Sze, ``Fastdepth: Fast monocular
  depth estimation on embedded systems,'' in \emph{2019 International
  Conference on Robotics and Automation (ICRA)}.\hskip 1em plus 0.5em minus
  0.4em\relax IEEE, 2019, pp. 6101--6108.

\bibitem{fu2018deep}
H.~Fu, M.~Gong, C.~Wang, K.~Batmanghelich, and D.~Tao, ``Deep ordinal
  regression network for monocular depth estimation,'' in \emph{Proceedings of
  the IEEE conference on computer vision and pattern recognition}, 2018, pp.
  2002--2011.

\bibitem{li2020height}
X.~Li, M.~Wang, and Y.~Fang, ``Height estimation from single aerial images
  using a deep ordinal regression network,'' \emph{IEEE Geoscience and Remote
  Sensing Letters}, 2020.

\bibitem{ghamisi2018img2dsm}
P.~Ghamisi and N.~Yokoya, ``Img2dsm: Height simulation from single imagery
  using conditional generative adversarial net,'' \emph{IEEE Geoscience and
  Remote Sensing Letters}, vol.~15, no.~5, pp. 794--798, 2018.

\bibitem{carvalho2018regression}
M.~Carvalho, B.~Le~Saux, P.~Trouv{\'e}-Peloux, A.~Almansa, and F.~Champagnat,
  ``On regression losses for deep depth estimation,'' in \emph{2018 25th IEEE
  International Conference on Image Processing (ICIP)}.\hskip 1em plus 0.5em
  minus 0.4em\relax IEEE, 2018, pp. 2915--2919.

\bibitem{srivastava2017joint}
S.~Srivastava, M.~Volpi, and D.~Tuia, ``Joint height estimation and semantic
  labeling of monocular aerial images with cnns,'' in \emph{2017 IEEE
  International Geoscience and Remote Sensing Symposium (IGARSS)}.\hskip 1em
  plus 0.5em minus 0.4em\relax IEEE, 2017, pp. 5173--5176.

\bibitem{zhang2019multi}
Y.~Zhang and X.~Chen, ``Multi-path fusion network for high-resolution height
  estimation from a single orthophoto,'' in \emph{2019 IEEE International
  Conference on Multimedia \& Expo Workshops (ICMEW)}.\hskip 1em plus 0.5em
  minus 0.4em\relax IEEE, 2019, pp. 186--191.

\bibitem{wang2015towards}
P.~Wang, X.~Shen, Z.~Lin, S.~Cohen, B.~Price, and A.~L. Yuille, ``Towards
  unified depth and semantic prediction from a single image,'' in
  \emph{Proceedings of the IEEE conference on computer vision and pattern
  recognition}, 2015, pp. 2800--2809.

\bibitem{lin1998building}
C.~Lin and R.~Nevatia, ``Building detection and description from a single
  intensity image,'' \emph{Computer vision and image understanding}, vol.~72,
  no.~2, pp. 101--121, 1998.

\bibitem{baatz1999object}
M.~Baatz, ``Object-oriented and multi-scale image analysis in semantic
  networks,'' in \emph{Proc. the 2nd International Symposium on
  Operationalization of Remote Sensing, Enschede, ITC, Aug. 1999}, 1999.

\bibitem{wang2005building}
Z.~Wang and W.~Liu, ``Building extraction from high resolution imagery based on
  multi-scale object oriented classification and probabilistic hough
  transform,'' in \emph{Proceedings of 2005 International Geoscience and Remote
  Sensing Symposium (IGARSS’05), Seoul, South Korea}, 2005, pp. 25--29.

\bibitem{mnih2013machine}
V.~Mnih, \emph{Machine learning for aerial image labeling}.\hskip 1em plus
  0.5em minus 0.4em\relax University of Toronto (Canada), 2013.

\bibitem{huang2016building}
Z.~Huang, G.~Cheng, H.~Wang, H.~Li, L.~Shi, and C.~Pan, ``Building extraction
  from multi-source remote sensing images via deep deconvolution neural
  networks,'' in \emph{2016 IEEE International Geoscience and Remote Sensing
  Symposium (IGARSS)}.\hskip 1em plus 0.5em minus 0.4em\relax Ieee, 2016, pp.
  1835--1838.

\bibitem{wu2018automatic}
G.~Wu, X.~Shao, Z.~Guo, Q.~Chen, W.~Yuan, X.~Shi, Y.~Xu, and R.~Shibasaki,
  ``Automatic building segmentation of aerial imagery using multi-constraint
  fully convolutional networks,'' \emph{Remote Sensing}, vol.~10, no.~3, p.
  407, 2018.

\bibitem{bulatov2014context}
D.~Bulatov, G.~H{\"a}ufel, J.~Meidow, M.~Pohl, P.~Solbrig, and P.~Wernerus,
  ``Context-based automatic reconstruction and texturing of 3d urban terrain
  for quick-response tasks,'' \emph{ISPRS Journal of Photogrammetry and Remote
  Sensing}, vol.~93, pp. 157--170, 2014.

\bibitem{yan2017hierarchical}
Y.~Yan, F.~Gao, S.~Deng, and N.~Su, ``A hierarchical building segmentation in
  digital surface models for 3d reconstruction,'' \emph{Sensors}, vol.~17,
  no.~2, p. 222, 2017.

\bibitem{li2016reconstructing}
M.~Li, L.~Nan, N.~Smith, and P.~Wonka, ``Reconstructing building mass models
  from uav images,'' \emph{Computers \& Graphics}, vol.~54, pp. 84--93, 2016.

\bibitem{yu2021automatic}
D.~Yu, S.~Ji, J.~Liu, and S.~Wei, ``Automatic 3d building reconstruction from
  multi-view aerial images with deep learning,'' \emph{ISPRS Journal of
  Photogrammetry and Remote Sensing}, vol. 171, pp. 155--170, 2021.

\bibitem{alidoost20192d}
F.~Alidoost, H.~Arefi, and F.~Tombari, ``2d image-to-3d model: Knowledge-based
  3d building reconstruction (3dbr) using single aerial images and
  convolutional neural networks (cnns),'' \emph{Remote Sensing}, vol.~11,
  no.~19, p. 2219, 2019.

\bibitem{bosch2019semantic}
M.~Bosch, K.~Foster, G.~Christie, S.~Wang, G.~D. Hager, and M.~Brown,
  ``Semantic stereo for incidental satellite images,'' in \emph{2019 IEEE
  Winter Conference on Applications of Computer Vision (WACV)}.\hskip 1em plus
  0.5em minus 0.4em\relax IEEE, 2019, pp. 1524--1532.

\bibitem{kim20023d}
J.~Kim, ``3d reconstruction from very high resolution satellite stereo and its
  application to object identification,'' \emph{The International Archives of
  the Photogrammetry, remote sensing and spatial information Sciences}, vol.~4,
  pp. 420--426, 2002.

\bibitem{kuschk2013model}
G.~Kuschk, ``Model-free dense stereo reconstruction for creating realistic 3d
  city models,'' in \emph{Joint Urban Remote Sensing Event 2013}.\hskip 1em
  plus 0.5em minus 0.4em\relax IEEE, 2013, pp. 202--205.

\bibitem{ozcanli2014automatic}
O.~C. Ozcanli, Y.~Dong, J.~L. Mundy, H.~Webb, R.~Hammoud, and T.~Victor,
  ``Automatic geo-location correction of satellite imagery,'' in
  \emph{Proceedings of the IEEE Conference on Computer Vision and Pattern
  Recognition Workshops}, 2014, pp. 307--314.

\bibitem{sun2021pbnet}
X.~Sun, P.~Wang, C.~Wang, Y.~Liu, and K.~Fu, ``Pbnet: Part-based convolutional
  neural network for complex composite object detection in remote sensing
  imagery,'' \emph{ISPRS Journal of Photogrammetry and Remote Sensing}, vol.
  173, pp. 50--65, 2021.

\bibitem{qi2017pointnet}
C.~R. Qi, H.~Su, K.~Mo, and L.~J. Guibas, ``Pointnet: Deep learning on point
  sets for 3d classification and segmentation,'' in \emph{Proceedings of the
  IEEE conference on computer vision and pattern recognition}, 2017, pp.
  652--660.

\bibitem{mao2022semantic}
Y.~Mao, X.~Sun, W.~Diao, K.~Chen, Z.~Guo, X.~Lu, and K.~Fu, ``Semantic
  segmentation for point cloud scenes via dilated graph feature aggregation and
  pyramid decoders,'' \emph{arXiv preprint arXiv:2204.04944}, 2022.

\bibitem{mao2022beyond}
Y.~Mao, K.~Chen, W.~Diao, X.~Sun, X.~Lu, K.~Fu, and M.~Weinmann, ``Beyond
  single receptive field: A receptive field fusion-and-stratification network
  for airborne laser scanning point cloud classification,'' \emph{ISPRS Journal
  of Photogrammetry and Remote Sensing}, vol. 188, pp. 45--61, 2022.

\bibitem{deng2021ccanet}
G.~Deng, Z.~Wu, C.~Wang, M.~Xu, and Y.~Zhong, ``Ccanet: Class-constraint
  coarse-to-fine attentional deep network for subdecimeter aerial image
  semantic segmentation,'' \emph{IEEE Transactions on Geoscience and Remote
  Sensing}, vol.~60, pp. 1--20, 2021.

\bibitem{vaswani2017attention}
A.~Vaswani, N.~Shazeer, N.~Parmar, J.~Uszkoreit, L.~Jones, A.~N. Gomez,
  {\L}.~Kaiser, and I.~Polosukhin, ``Attention is all you need,''
  \emph{Advances in neural information processing systems}, vol.~30, 2017.

\bibitem{mou2020relation}
L.~Mou, Y.~Hua, and X.~X. Zhu, ``Relation matters: Relational context-aware
  fully convolutional network for semantic segmentation of high-resolution
  aerial images,'' \emph{IEEE Transactions on Geoscience and Remote Sensing},
  vol.~58, no.~11, pp. 7557--7569, 2020.

\bibitem{wei2020oriented}
H.~Wei, Y.~Zhang, Z.~Chang, H.~Li, H.~Wang, and X.~Sun, ``Oriented objects as
  pairs of middle lines,'' \emph{ISPRS Journal of Photogrammetry and Remote
  Sensing}, vol. 169, pp. 268--279, 2020.

\bibitem{wei2020x}
H.~Wei, Y.~Zhang, B.~Wang, Y.~Yang, H.~Li, and H.~Wang, ``X-linenet: Detecting
  aircraft in remote sensing images by a pair of intersecting line segments,''
  \emph{IEEE Transactions on Geoscience and Remote Sensing}, vol.~59, no.~2,
  pp. 1645--1659, 2020.

\bibitem{wang2016efficient}
K.~Wang, C.~Stutts, E.~Dunn, and J.-M. Frahm, ``Efficient joint stereo
  estimation and land usage classification for multiview satellite data,'' in
  \emph{2016 IEEE Winter Conference on Applications of Computer Vision
  (WACV)}.\hskip 1em plus 0.5em minus 0.4em\relax IEEE, 2016, pp. 1--9.

\bibitem{wu2014building}
B.~Wu, X.~Sun, Q.~Wu, M.~Yan, H.~Wang, and K.~Fu, ``Building reconstruction
  from high-resolution multiview aerial imagery,'' \emph{IEEE Geoscience and
  Remote Sensing Letters}, vol.~12, no.~4, pp. 855--859, 2014.

\bibitem{hepp2018plan3d}
B.~Hepp, M.~Nie{\ss}ner, and O.~Hilliges, ``Plan3d: Viewpoint and trajectory
  optimization for aerial multi-view stereo reconstruction,'' \emph{ACM
  Transactions on Graphics (TOG)}, vol.~38, no.~1, pp. 1--17, 2018.

\bibitem{Cramer_2010}
M.~Cramer, ``{The DGPF-test on digital airborne camera evaluation -- Overview
  and test design},'' \emph{{PFG Photogrammetrie -- Fernerkundung --
  Geoinformation}}, vol. {2 / 2010}, pp. 73--82, 2010.

\bibitem{kim2013simulation}
S.~Kim, I.~Lee, and Y.~J. Kwon, ``Simulation of a geiger-mode imaging ladar
  system for performance assessment,'' \emph{sensors}, vol.~13, no.~7, pp.
  8461--8489, 2013.

\bibitem{Rottensteiner_et_al_2012}
F.~Rottensteiner, G.~Sohn, J.~Jung, M.~Gerke, C.~Baillard, S.~Benitez, and
  U.~Breitkopf, ``{The ISPRS benchmark on urban object classification and 3D
  building reconstruction},'' \emph{{ISPRS Annals of the Photogrammetry, Remote
  Sensing and Spatial Information Sciences}}, vol. {I-3}, pp. 293--298, 2012.

\bibitem{niemeyer2014contextual}
J.~Niemeyer, F.~Rottensteiner, and U.~Soergel, ``Contextual classification of
  lidar data and building object detection in urban areas,'' \emph{ISPRS
  Journal of Photogrammetry and Remote Sensing}, vol.~87, pp. 152--165, 2014.

\bibitem{ye2020lasdu}
Z.~Ye, Y.~Xu, R.~Huang, X.~Tong, X.~Li, X.~Liu, K.~Luan, L.~Hoegner, and
  U.~Stilla, ``Lasdu: A large-scale aerial lidar dataset for semantic labeling
  in dense urban areas,'' \emph{ISPRS International Journal of
  Geo-Information}, vol.~9, no.~7, p. 450, 2020.

\bibitem{le20192019}
B.~Le~Saux, N.~Yokoya, R.~H{\"a}nsch, and M.~Brown, ``2019 ieee grss data
  fusion contest: large-scale semantic 3d reconstruction,'' \emph{IEEE
  Geoscience and Remote Sensing Magazine (GRSM)}, vol.~7, no.~4, pp. 33--36,
  2019.

\bibitem{li2020semantic}
X.~Li, A.~You, Z.~Zhu, H.~Zhao, M.~Yang, K.~Yang, S.~Tan, and Y.~Tong,
  ``Semantic flow for fast and accurate scene parsing,'' in \emph{European
  Conference on Computer Vision}.\hskip 1em plus 0.5em minus 0.4em\relax
  Springer, 2020, pp. 775--793.

\bibitem{he2016deep}
K.~He, X.~Zhang, S.~Ren, and J.~Sun, ``Deep residual learning for image
  recognition,'' in \emph{Proceedings of the IEEE conference on computer vision
  and pattern recognition}, 2016, pp. 770--778.

\bibitem{chen2018encoder}
L.-C. Chen, Y.~Zhu, G.~Papandreou, F.~Schroff, and H.~Adam, ``Encoder-decoder
  with atrous separable convolution for semantic image segmentation,'' in
  \emph{Proceedings of the European conference on computer vision (ECCV)},
  2018, pp. 801--818.

\bibitem{zwald2012berhu}
L.~Zwald and S.~Lambert-Lacroix, ``The berhu penalty and the grouped effect,''
  \emph{arXiv preprint arXiv:1207.6868}, 2012.

\bibitem{kazhdan2006poisson}
M.~Kazhdan, M.~Bolitho, and H.~Hoppe, ``Poisson surface reconstruction,'' in
  \emph{Proceedings of the fourth Eurographics symposium on Geometry
  processing}, vol.~7, 2006.

\bibitem{ba2016layer}
J.~L. Ba, J.~R. Kiros, and G.~E. Hinton, ``Layer normalization,'' \emph{arXiv
  preprint arXiv:1607.06450}, 2016.

\bibitem{jaderberg2015spatial}
M.~Jaderberg, K.~Simonyan, A.~Zisserman \emph{et~al.}, ``Spatial transformer
  networks,'' \emph{Advances in neural information processing systems},
  vol.~28, 2015.

\bibitem{zhao2017pyramid}
H.~Zhao, J.~Shi, X.~Qi, X.~Wang, and J.~Jia, ``Pyramid scene parsing network,''
  in \emph{Proceedings of the IEEE conference on computer vision and pattern
  recognition}, 2017, pp. 2881--2890.

\bibitem{3dfier}
H.~Ledoux, F.~Biljecki, B.~Dukai, K.~Kumar, R.~Peters, J.~Stoter, and
  T.~Commandeur, ``3dfier: automatic reconstruction of 3d city models,''
  \emph{Journal of Open Source Software}, vol.~6, no.~57, p. 2866, 2021.

\bibitem{hajek2016principles}
P.~Hajek, K.~Jedli{\v{c}}ka, and V.~{\v{C}}ada, ``Principles of cartographic
  design for 3d maps--focused on urban areas,'' in \emph{6th International
  Conference on Cartography and GIS Proceedings}, vol.~1, 2016, pp. 297--307.

\bibitem{mittal2019vision}
M.~Mittal, R.~Mohan, W.~Burgard, and A.~Valada, ``Vision-based autonomous uav
  navigation and landing for urban search and rescue,'' \emph{arXiv preprint
  arXiv:1906.01304}, 2019.

\bibitem{vaihingen.org}
``Isprs.2d semantic labeling contest-vaihingen,'' \emph{[Online]. Available:
  http://www2.isprs.org/commissions/comm3/wg4/2d-sem-label-vaihingen.html}.

\end{thebibliography}

\end{document}